\documentclass{article} 
\usepackage{iclr2020_conference,times}

\usepackage{hyperref}
\usepackage{times}
\usepackage{epsfig}
\usepackage{graphicx}
\usepackage{amsmath}
\usepackage{amssymb}

\usepackage{wrapfig}
\usepackage[utf8]{inputenc} 
\usepackage[T1]{fontenc}    
\usepackage{url}            
\usepackage{booktabs}       
\usepackage{amsfonts}       
\usepackage{nicefrac}       
\usepackage{microtype}      
\usepackage{xspace}
\usepackage{caption}
\usepackage{subcaption}
\usepackage[ruled, vlined]{algorithm2e}
\usepackage{cprotect}
\usepackage{hhline}
\usepackage{array}
\definecolor{lightpurple}{rgb}{178,145,226}
\definecolor{lightblue}{rgb}{0,128,255}
\definecolor{yellow}{rgb}{1,1,0}

\renewcommand{\paragraph}[1]{\vspace{0.0ex}\textbf{#1}}
\newcommand{\ie}{i.e.\xspace}
\newcommand{\eg}{e.g.\xspace}

\newcommand*{\rowstyle}[1]{
	\gdef\@rowstyle{#1}
	\@rowstyle\ignorespaces%
}
\newcolumntype{=}{
	>{\gdef\@rowstyle{}}
}
\newcolumntype{+}{
	>{\@rowstyle}
}

\usepackage{amssymb}
\usepackage{amsmath,amsfonts}
\usepackage{amsopn}
\usepackage{bm} 
\usepackage{multirow}

\newcommand{\field}[1]{\mathbb{#1}}
\newcommand{\R}{\field{R}} 

\newcommand{\ProbOpr}[1]{\mathbb{#1}}

\newcommand{\expect}[2]{%
\ifthenelse{\equal{#2}{}}{\ProbOpr{E}_{#1}}
{\ifthenelse{\equal{#1}{}}{\ProbOpr{E}\left[#2\right]}{\ProbOpr{E}_{#1}\left[#2\right]}}} 
\newcommand{\var}[2]{%
\ifthenelse{\equal{#2}{}}{\ProbOpr{VAR}_{#1}}
{\ifthenelse{\equal{#1}{}}{\ProbOpr{VAR}\left[#2\right]}{\ProbOpr{VAR}_{#1}\left[#2\right]}}} 

\DeclareMathOperator{\argmin}{arg\,min}
\DeclareMathOperator{\softmax}{softmax}

\newcommand{\eat}[1]{}
\newcommand{\method}[1]{\textsc{#1}}

\newcommand{\GDC}{\method{GDC}\xspace}
\newcommand{\SDN}{\method{SDN}\xspace}

\newcommand{\vPIXOR}{\method{PIXOR$^\star$}\xspace}
\newcommand{\PIXOR}{\method{PIXOR}\xspace}
\newcommand{\PRCNN}{\method{P-RCNN}\xspace}
\newcommand{\SRCNN}{\method{S-RCNN}\xspace}
\newcommand{\PL}{\method{pseudo-LiDAR}\xspace}
\newcommand{\AVOD}{\method{AVOD}\xspace}
\newcommand{\Frustum}{\method{F-PointNet}\xspace}

\newcommand{\DOP}{\method{3DOP}\xspace}

\newcommand{\MLFstereo}{\method{MLF-stereo}\xspace}

\newcommand{\APBEV}{AP$_\text{BEV}$\xspace}
\newcommand{\AP}{AP$_\text{3D}$\xspace}

\newcommand{\PSMNet}{\method{PSMNet}\xspace}
\newcommand{\PSMNetpd}{\method{PSMNet}\xspace}

\newcommand{\PnP}{\method{PnP}\xspace}

\ificlrfinal\fi

\title{Pseudo-LiDAR++: \\Accurate Depth for 3D Object Detection in\\ Autonomous Driving}

\author{Yurong You$^{*1}$, Yan Wang$^{*1}$, Wei-Lun Chao\thanks{\hspace{1pt}Equal contributions}\hspace{4pt}$^2$, Divyansh Garg$^1$, Geoff Pleiss$^1$,\\
	\textbf{Bharath Hariharan$^1$, Mark Campbell$^1$, {\normalfont and} Kilian Q. Weinberger$^1$}\\
	$^1$Cornell University, Ithaca, NY \hspace{20pt}$^2$The Ohio State University, Columbus, OH\\
	{\tt\small \{yy785, yw763, dg595, gp346, bh497, mc288, kqw4\}@cornell.edu}\\
	{\tt\small chao.209@osu.edu}
}

\iclrfinalcopy 
\begin{document}
\maketitle

\begin{abstract}
Detecting objects such as cars and pedestrians in 3D plays an indispensable role in autonomous driving. Existing approaches largely rely on expensive LiDAR sensors for accurate depth information. While recently pseudo-LiDAR has been introduced as a promising alternative, at a much lower cost based solely on stereo images, there is still a notable performance gap. 
In this paper we provide substantial advances to the pseudo-LiDAR framework through improvements in stereo depth estimation. 
Concretely, we adapt the stereo network architecture and loss function to be more aligned with accurate depth estimation of faraway objects --- currently the primary weakness of pseudo-LiDAR. 
Further, we explore the idea to leverage cheaper but extremely sparse LiDAR sensors, which alone provide insufficient information for 3D detection, to de-bias our depth estimation. We propose a depth-propagation algorithm, guided by the initial depth estimates, to diffuse these few exact measurements across the entire depth map. 
We show on the KITTI object detection benchmark that our combined approach yields substantial improvements in depth estimation and stereo-based 3D object detection --- outperforming the previous state-of-the-art detection accuracy for faraway objects by $40\%$. Our code is available at \url{https://github.com/mileyan/Pseudo_Lidar_V2}.
\end{abstract}

 
\section{Introduction}
 
\label{sec:intro}
\begin{wrapfigure}{r}{0.5\textwidth}
	\centering
	\vspace{-11pt}
	\includegraphics[width=.95\linewidth]{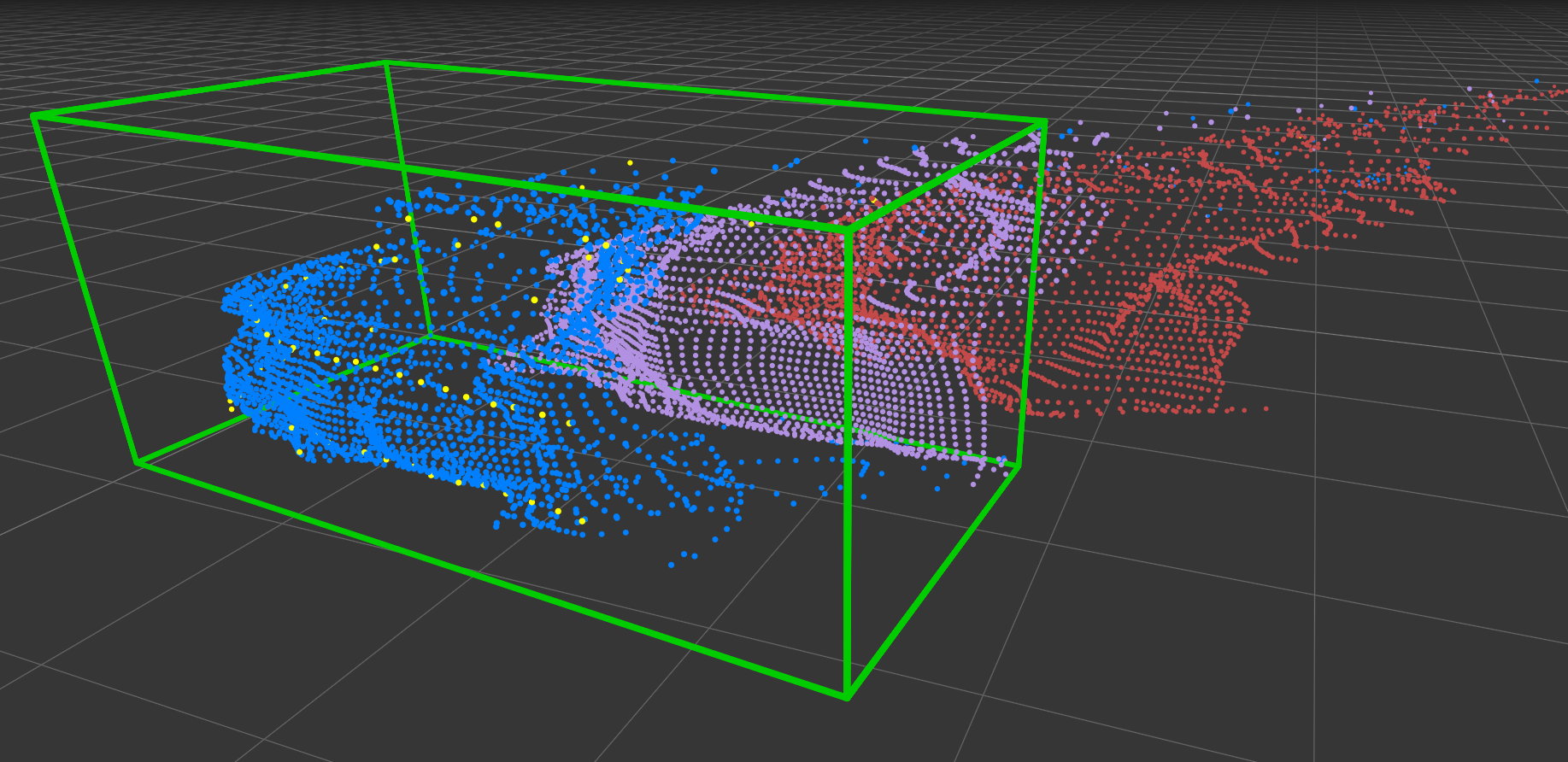}
	\vskip -5pt
	\caption{\small \textbf{An illustration of our proposed depth estimation and correction method.} The {\color{green}green} box is the ground truth location of the car in the KITTI dataset. The {\color{red}red} points are obtained with a stereo disparity network. {\color{violet}Purple} points, obtained with our stereo depth network (SDN), are much closer to the truth. After depth propagation ({\color{blue}blue} points) with a few ({\color{yellow}yellow}) LiDAR measurements the car is squarely inside the green box. (One floor square is 1m$\times$1m.)  \label{fig:redpurpleblue}}
	\vspace{-20pt}
\end{wrapfigure}
Safe driving in autonomous cars requires accurate 3D detection and localization of cars, pedestrians and other objects. This in turn requires accurate depth information, which can be obtained from LiDAR (Light Detection And Ranging) sensors. Although highly precise and reliable, LiDAR sensors are notoriously expensive:  a 64-beam model can cost around \$75,000 (USD)\footnote{The information is obtained from the automotive LiDAR market report:~\url{http://www.woodsidecap.com/wp-content/uploads/2018/04/Yole_WCP-LiDAR-Report_April-2018-FINAL.pdf}}.
The alternative is to measure depth through inexpensive commodity cameras.
However, in spite of recent dramatic progress in stereo-based 3D object detection brought by pseudo-LiDAR~\citep{pseudoLiDAR}, a significant performance gap remains especially for faraway objects (which we want to detect early to allow time for reaction). 
The trade-off between affordability and safety creates an ethical dilemma. 

In this paper we propose a possible solution to this remaining challenge that combines insights from both perspectives. We observe that the higher 3D object localization error of stereo-based systems, compared to LiDAR-based ones, stems entirely from the higher error in depth estimation (after the 3D point cloud is obtained the two approaches are identical~\citep{pseudoLiDAR}). Importantly, this error is not random but \emph{systematic}: we observe that stereo methods do indeed \emph{detect} objects with high reliability, yet they estimate the depth of the \emph{entire} object as either too far or too close. See \autoref{fig:redpurpleblue} for an illustration: the red stereo points capture the car but are shifted by about 2m completely outside the ground-truth location (green box). 
If we can \emph{de-bias} these depth estimates it should be possible to obtain accurate 3D localization even for distant objects without exorbitant costs. 

We start by revisiting the depth estimation routine embedded at the heart of state-of-the-art stereo-based 3D detection approach~\citep{pseudoLiDAR}.
A major contributor to the systematic depth bias comes from the fact that depth is typically not computed directly. Instead, one first estimates the \emph{disparity} --- the horizontal shift of a pixel between the left and right images --- and then \emph{inverts} it to obtain pixel-wise depth. 
While the use of deep neural networks has largely improved disparity estimation~\citep{chang2018pyramid,cheng2018depth,mayer2016large,wang2018anytime}, designing and learning the networks to optimize the accuracy of \emph{disparity estimation} simply over-emphasizes nearby objects due to the reciprocal transformation.
For instance, a unit disparity error (in pixels) for a 5-meter-away object means a 10cm error in depth: the length of a side mirror. The same disparity error for a 50-meter-away object, however, becomes a 5.8m error in depth: the length of an entire car.
Penalizing both errors equally means that the network spends more time correcting subtle errors on nearby objects than gross errors on faraway objects, resulting in degraded depth estimates and ultimately poor detection and localization for faraway objects. 
We thus propose to adapt the stereo network architecture and loss function for direct depth estimation.
Concretely, the cost volume that fuses the left-right images and the subsequent 3D convolutions are the key components in stereo networks. Taking the central assumption of convolutions --- all neighborhoods can be operated in an identical manner --- we propose to construct the cost volume on the grid of depth rather than disparity, enabling 3D convolutions and the loss function to perform exactly on the right scale for depth estimation.
We refer to our network as stereo depth network (\SDN). See~\autoref{fig:redpurpleblue}  for a comparison of 3D points obtained with SDN (purple) and disparity estimation (red). 

Although our \SDN{} improves the depth estimates significantly, stereo images are still inherently 2D and it is unclear if they can ever match the accuracy and reliability of a true 3D LiDAR sensor. Although LiDAR sensors with 32 or 64 beams are expensive, LiDAR sensors with only 4 beams are two orders of magnitude cheaper\footnote{The Ibeo Wide Angle Scanning (ScaLa) sensor with 4 beams costs \$600 (USD). In this paper we simulate the 4-beam LiDAR signal on KITTI benchmark~\citep{geiger2012we} by sparsifying the original 64-beam signal.} and thus easily affordable. 
The 4 laser beams are very sparse and ill-suited to capture 3D object shapes by themselves, but if paired with stereo images they become the ideal tool to de-bias our dense stereo depth estimates: a single high-precision laser beam may inform us how to correct the depth of an entire car or pedestrian in its path. 
To this end, we present a novel depth-propagation algorithm, inspired by graph-based manifold learning~\citep{WeinbergerPS05,roweis2000nonlinear,xiaojin2002learning}. In a nutshell, we connect our estimated 3D stereo point cloud locally by a nearest neighbor graph, such that points corresponding to the same object will share many local paths with each other. We match the few but exact LiDAR measurements first with pixels (irrespective of depth) and then with their corresponding 3D points to obtain accurate depth estimates for several nodes in the graph. Finally, we propagate this exact depth information along the graph using a label diffusion mechanism 
--- resulting in a \emph{dense and accurate depth map} at \emph{negligible cost}. In \autoref{fig:redpurpleblue} we see that the few (yellow) LiDAR measurements are sufficient to position almost all final (blue) points of the entire car within the green ground truth box. 

We conduct extensive empirical studies of our approaches on the KITTI object detection benchmark~\citep{geiger2012we,geiger2013vision} and achieve remarkable results. With solely stereo images, we outperform the previous state of the art~\citep{pseudoLiDAR} by $10\%$. Further adding a cheap 4-beam LiDAR brings another $27\%$ relative improvement --- on some metrics, our approach is nearly on par with those based on a 64-beam LiDAR but can potentially save $95\%$ in cost.


\section{Background}

\label{sec:background}
\noindent\textbf{3D object detection.}
Most work on 3D object detection operates on 3D point clouds from LiDAR as input~\citep{li20173d,li2016vehicle,meyer2019lasernet,yang2018hdnet,du2018general,shi2019pointrcnn,engelcke2017vote3deep,yan2018second,lang2019pointpillars}. Frustum PointNet~\citep{qi2018frustum} applies PointNet~\citep{qi2017pointnet,qi2017pointnet++} to the points directly, while Voxelnet~\citep{zhou2018voxelnet} quantizes them into 3D grids. For street scenes, several work finds that processing points from the bird's-eye view can already capture object contours and locations~\citep{chen2017multi,yang2018pixor,ku2018joint}. Images have also been used, but mainly to supplement LiDAR~\citep{meyer2019sensor,xu2018pointfusion,liang2018deep,chen2017multi,ku2018joint}.
Early work based solely on images --- mostly built on the 2D frontal-view detection pipeline~\citep{ren2015faster,he2017mask,lin2017feature} --- fell far behind in localizing objects in 3D~\citep{li2019gs3d,xiang2015data,xiang2017subcategory,chabot2017deep,mousavian20173d,chen20153d,xu2018multi,chen2016monocular,pham2017robust,chen20183d}\footnote{Recently, \cite{srivastava2019learning} proposed to lift 2D monocular images to 3D representations (e.g., bird's-eye view (BEV) images) and achieved promising monocular-based 3D object detection results.}.

\noindent\textbf{Pseudo-LiDAR.}
This gap has been reduced significantly recently with the introduction of the pseudo-LiDAR framework proposed in~\citep{pseudoLiDAR}.
This framework applies a drastically different approach from previous image-based 3D object detectors. Instead of directly detecting the 3D bounding boxes from the frontal view of a scene, pseudo-LiDAR begins with image-based depth estimation, predicting the depth $Z(u,v)$ of each image pixel $(u,v)$. The resulting depth map $Z$ is then back-projected into a 3D point cloud: a pixel $(u, v)$ will be transformed to $(x,y,z)$ in 3D by
\begin{align}
z = Z(u, v), \hspace{20pt}  x = \frac{(u - c_U)\times z}{f_U}, \hspace{20pt} y = \frac{(v - c_V)\times z}{f_V}, \label{eq_3D}
\end{align}
where $(c_U, c_V)$ is the camera center and $f_U$ and $f_V$ are the horizontal and vertical focal length. The 3D point cloud is then treated exactly as LiDAR signal --- any LiDAR-based 3D detector can be applied seamlessly. By taking the state-of-the-art algorithms from both ends~\citep{chang2018pyramid, ku2018joint, qi2018frustum}, pseudo-LiDAR obtains the highest image-based performance on the KITTI object detection benchmark~\citep{geiger2012we,geiger2013vision}.
Our work builds upon this framework.

\noindent\textbf{Stereo disparity estimation.}
Pseudo-LiDAR relies heavily on the quality of depth estimation.
Essentially, if the estimated pixel depths match those provided by LiDAR, pseudo-LiDAR with any LiDAR-based detector should be able to achieve the same performance as that obtained by applying the same detector to the LiDAR signal.
According to~\citep{pseudoLiDAR}, depth estimation from stereo pairs of images~\citep{mayer2016large,yamaguchi2014efficient,chang2018pyramid} are more accurate than that from monocular (\ie, single) images~\citep{fu2018deep,godard2017unsupervised} for 3D object detection. We therefore focus on stereo depth estimation, which is routinely obtained from estimating disparity between images.

A disparity estimation algorithm takes a pair of left-right images $I_l$ and $I_r$ as input, captured from a pair of cameras with a horizontal offset (i.e., baseline) $b$. Without loss of generality, we assume that the algorithm treats the left image, $I_l$, as reference and outputs a disparity map $D$ recording the horizontal disparity to $I_r$ for each pixel $(u, v)$. Ideally, $I_l(u, v)$ and $I_r(u, v + D(u, v))$ will picture the same 3D location.
We can therefore derive the depth map $Z$ via the following transform,
\begin{align}
Z(u, v) = \frac{f_U \times b}{D(u, v)} \quad (f_U\text{: horizontal focal length}). \label{eq_disp_depth}
\end{align}
A common pipeline of disparity estimation is to first construct a 4D disparity cost volume $C_\text{disp}$, in which $C_\text{disp}(u,v,d,:)$ is a feature vector that captures the pixel difference between $I_l(u,v)$ and $I_r(u,v + d)$. It then estimates the disparity $D(u, v)$ for each pixel $(u, v)$ according to the cost volume $C_\text{disp}$. One basic algorithm is to build a 3D cost volume with $C_\text{disp}(u,v,d) = \|I_l(u,v) - I_r(u,v + d)\|_2$ and determine $D(u,v)$ as $\argmin_d C_\text{disp}(u,v,d)$. Advanced algorithms exploit more robust features in constructing $C_\text{disp}$ and perform structured prediction for $D$. In what follows, we give an introduction of PSMNet~\citep{chang2018pyramid}, a state-of-the-art algorithm used in~\citep{pseudoLiDAR}.

PSMNet begins with extracting deep feature maps $h_l$ and $h_r$ from $I_l$ and $I_r$, respectively. It then constructs $C_\text{disp}(u,v,d,:)$ by concatenating features of $h_l(u,v)$ and $h_r(u,v+d)$, followed by layers of 3D convolutions. The resulting 3D tensor $S_\text{disp}$, with the feature channel size ending up being one, is then used to derive the pixel disparity via the following weighted combination,
\begin{align}
D(u, v) = \sum_d \softmax(-S_\text{disp}(u,v,d))\times d, \label{eq_disp_pred}
\end{align}
where $\softmax$ is performed along the 3\textsuperscript{rd} dimension of $S_\text{disp}$. PSMNet can be learned end-to-end, including the image feature extractor and 3D convolution kernels, to minimize the disparity error
\begin{align}
\sum_{(u,v)\in \mathcal{A}} \ell(D(u,v) - D^\star(u,v)), \label{eq_disp_loss}
\end{align}
where $\ell$ is the smooth L1 loss, $D^\star$ is the ground truth map, and $\mathcal{A}$ contains pixels with ground truths.


\section{Stereo Depth Network (SDN)}

\label{sec:approach}
\begin{wrapfigure}{r}{0.3\textwidth}
	\vspace{-13pt}
	\centering
	\includegraphics[width=\linewidth]{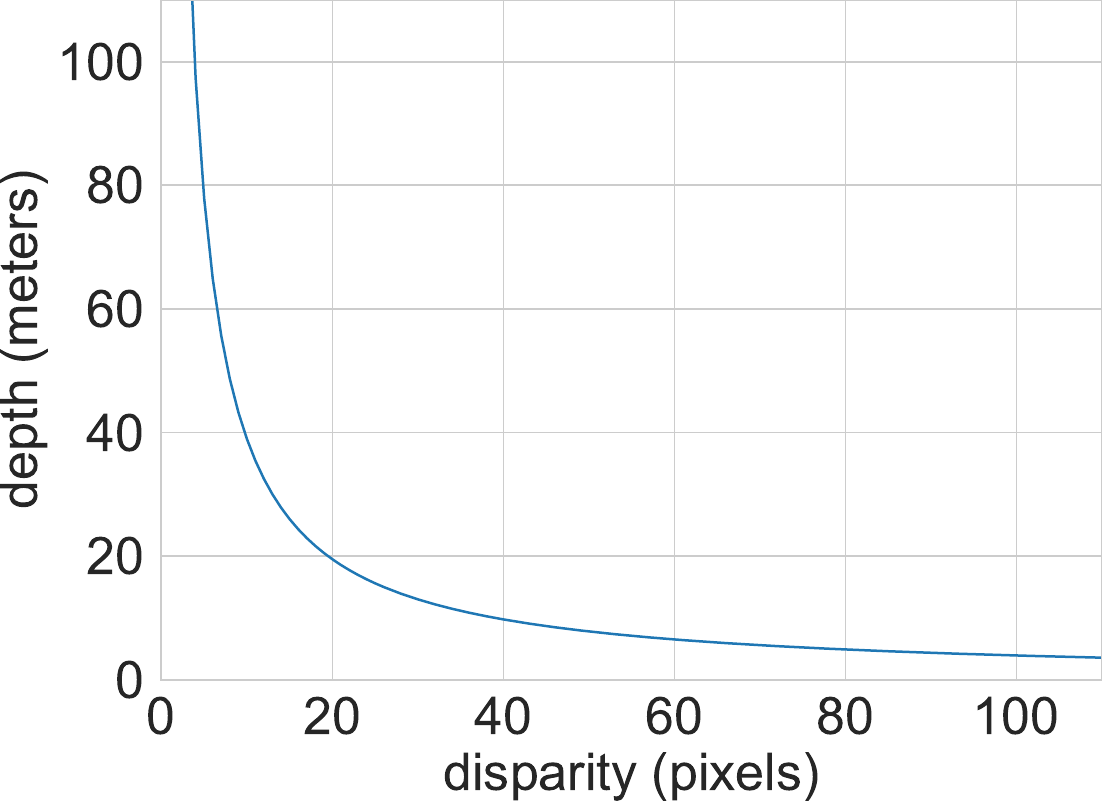}
	\vskip -8pt
	\caption{\small \textbf{The disparity-to-depth transform.} We set $f_U=721$ (in pixels) and $b=0.54$ (in meters) in \autoref{eq_disp_depth}, which are the typical values used in the KITTI dataset.}
	\label{fig:depth_disparity}
	\vspace{-10pt}
\end{wrapfigure}
A stereo network designed and  learned to minimize the disparity error (cf. \autoref{eq_disp_loss}) may over-emphasize nearby objects with smaller depths and therefore perform poorly in estimating depths for faraway objects.
To see this, note that \autoref{eq_disp_depth} implies that
for a given error in disparity $\delta D$, the error in depth $\delta Z$ increases \emph{quadratically} with depth:
\begin{equation}
Z \propto \frac{1}{D} \Rightarrow \delta Z \propto \frac{1}{D^2} \delta D \Rightarrow \delta Z \propto Z^2 \delta D.
\end{equation}
The middle term is obtained by differentiating $Z(D)$  w.r.t. $D$.
In particular, using the settings on the KITTI dataset~\citep{geiger2012we,geiger2013vision}, a single pixel error in disparity implies only a 0.1m error in depth at a depth of 5 meters, but a 5.8m error at a depth of 50 meters. See~\autoref{fig:depth_disparity} for a mapping from disparity to depth.

\noindent\textbf{Depth Loss.} We propose two changes to adapt stereo networks for direct depth estimation. First, we learn stereo networks to directly optimize the depth loss
\begin{align}
\sum_{(u,v)\in \mathcal{A}} \ell(Z(u,v) - Z^\star(u,v)). \label{eq_depth_loss}
\end{align}
$Z$ and $Z^\star$ can be obtained from $D$ and $D^\star$ using \autoref{eq_disp_depth}. The change from the disparity loss to the depth loss corrects the disproportionally strong emphasis on tiny depth errors of nearby objects --- a necessary but still insufficient change to overcome the problems of disparity estimation.

\begin{figure}
    \begin{minipage}[b]{.48\textwidth}
    \centering
	\includegraphics[width=\textwidth]{figures/combined}
	\caption{\small \textbf{Disparity cost volume (left) vs. depth cost volume (right).} The figure shows the 3D points obtained from LiDAR ({\color{yellow}yellow}) and stereo ({\color{violet}purple}) corresponding to a car in KITTI, seen from the bird's-eye view (BEV).
	Points from the disparity cost volume are stretched out and noisy; while points from the depth cost volume capture the car contour faithfully.}
	\label{fig:cost}
    \end{minipage}
    \hfill
    \begin{minipage}[b]{.48\textwidth}
   	\centering
	\includegraphics[width=0.75\textwidth]{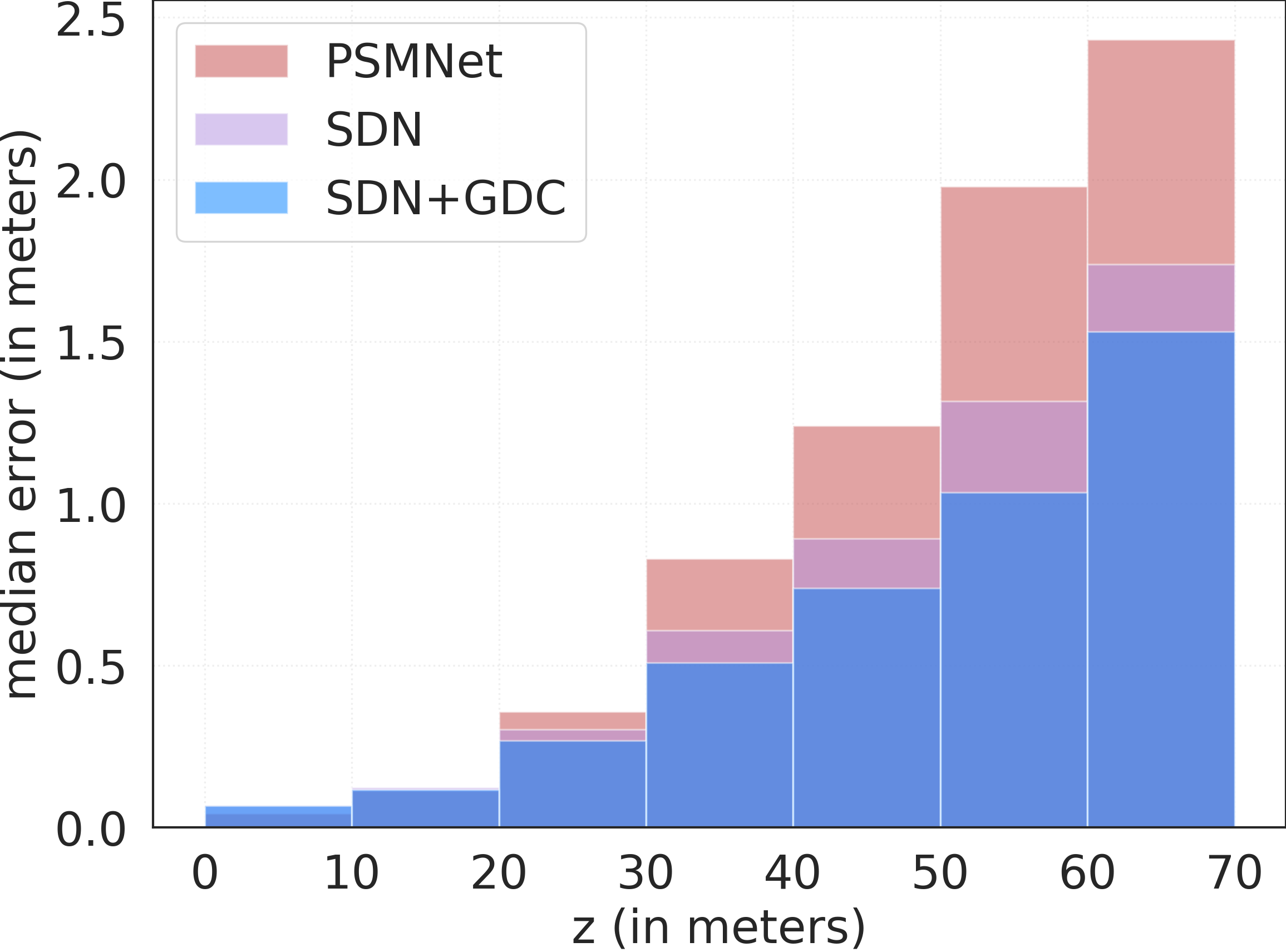}
	\vskip -8pt
	\caption{\small \textbf{Depth estimation errors.} We compare depth estimation error on 3,769 KITTI validation images, taking 64-beam LiDAR depths as ground truths. We separate pixels according to their true depths (z). See the text and appendix for details.}
	\label{fig:depth_comp}
    \end{minipage}
    \vspace{-7pt}
\end{figure}
\noindent\textbf{Depth Cost Volume.} To facilitate accurate depth learning (rather than disparity) we need to further address the internals of the depth estimation pipeline.
A crucial source of error is the 3D convolutions within the 4D disparity cost volume, where the  same  kernels are applied for the entire cost volume.
This is highly problematic as it implicitly assumes that the effect of a convolution is homogeneous throughout --- which is clearly violated by the reciprocal depth to disparity relation (\autoref{fig:depth_disparity}).
For example, it may be completely appropriate to locally smooth two neighboring pixels with disparity 85 and 86 (changing the depth by a few cm to smooth out a surface), whereas applying the same kernel for two pixels with disparity 5 and 6 could easily move the 3D points by 10m or more.

\begin{figure*}[t]
	\centering
	\includegraphics[width=0.95\linewidth]{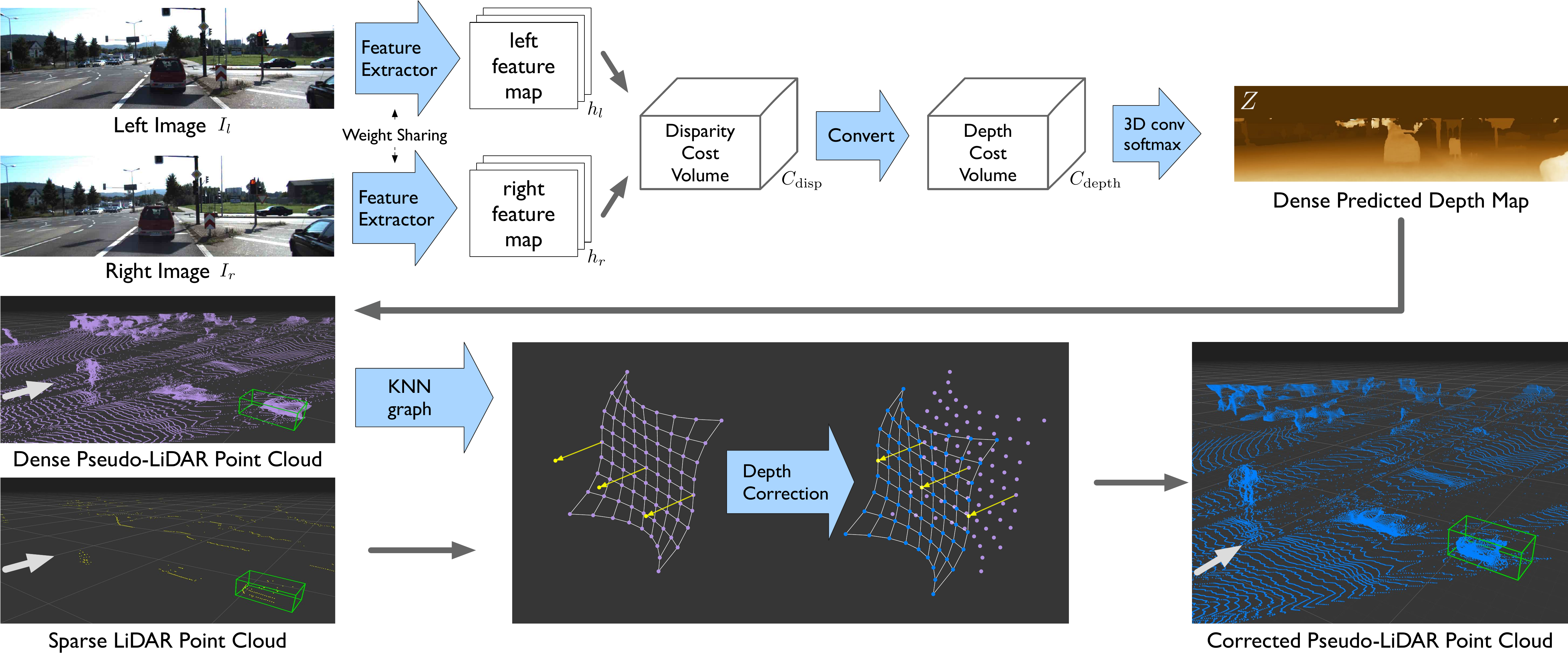}
	\vskip -5pt
	\caption{\small \textbf{The whole pipeline of improved stereo depth estimation:} (top) the stereo depth network (\SDN) constructs a depth cost volume from left-right images and is optimized for direct depth estimation; (bottom) the graph-based depth correction algorithm (\GDC) refines the depth map by leveraging sparser LiDAR signal. The gray arrows indicates the observer's view point. We superimpose the (green) ground-truth 3D box of a car, the same one in \autoref{fig:redpurpleblue}. The corrected points (blue; bottom right) are perfectly located inside the ground truth box.}
	\label{fig:PL++}
	\vspace{-10pt}
\end{figure*}

Taking this insight and the central assumption of convolutions --- all neighborhoods can be operated upon in an identical manner --- into account, we propose to instead construct the depth cost volume $C_\text{depth}$, in which $C_\text{depth}(u, v, z, :)$ will encode features describing how likely the depth $Z(u,v)$ of pixel $(u,v)$ is $z$. The subsequent 3D convolutions will then operate on the grid of depth, rather than disparity, affecting neighboring depths identically, independent of their location.  The resulting 3D tensor $S_\text{depth}$ is then used to predict the pixel depth similar to \autoref{eq_disp_pred}
\begin{equation}
Z(u, v) = \sum_z \softmax(-S_\text{depth}(u,v,z))\times z. \nonumber
\end{equation}
We construct the new depth volume, $C_\text{depth}$, based on the intuition that $C_\text{depth}(u,v,z,:)$ and $C_\text{disp}\left(u,v,\cfrac{f_U\times b}{z},:\right)$ should lead to equivalent ``cost''. To this end, we apply a bilinear interpolation to construct $C_\text{depth}$ from $C_\text{disp}$ using the depth-to-disparity transform in \autoref{eq_disp_depth}.
Specifically, we consider disparity in the range of $[0, 191]$ following PSMNet~\citep{chang2018pyramid}, and consider depth in the range of $[1\text{m}, 80\text{m}]$ and set the grid of depth in $C_\text{depth}$ to be 1m.
\autoref{fig:PL++} (top) depicts our stereo depth network (\SDN) pipeline.
Crucially, all convolution operations are operated on $C_\text{depth}$ exclusively.
\autoref{fig:depth_comp} compares the median values of absolute depth estimation errors using the disparity cost volume (i.e., PSMNet) and the depth cost volume (\SDN{}) (see \autoref{sec::supp_stereo_error} for detailed numbers). As expected, for faraway depth, \SDN{} leads to drastically smaller errors with only marginal increases in the very near range (which disparity based methods over-optimize). See the appendix for the detailed setup and more discussions.

\section{Depth Correction}
\label{sec:depth_correction}
\vspace{-1.5pt}
Our \SDN significantly improves depth estimation and more precisely renders the object contours (see \autoref{fig:cost}). However, there is a fundamental limitation in stereo because of the discrete nature of pixels: the disparity, being the difference in the horizontal coordinate between corresponding pixels, has to be \emph{quantized} at the level of individual pixels while the depth is \emph{continuous}.
Although the quantization error can be alleviated with higher resolution images, the computational depth prediction cost scales \emph{cubically} with resolution--- pushing the limits of GPUs in autonomous vehicles.

We therefore explore a hybrid approach by leveraging a cheap LiDAR with extremely sparse (\eg, 4 beams) but accurate depth measurements to \emph{correct} this bias.
We note that such sensors are too \emph{sparse} to capture object shapes and cannot be used alone for detection. However, by projecting the LiDAR points into the image plane we obtain exact depths on a small portion of ``landmark'' pixels.

We present a graph-based depth correction (\GDC{}) algorithm that effectively combines the \emph{dense} stereo depth that has rendered object shapes and the \emph{sparse} accurate LiDAR measurements. Conceptually, we expect the corrected depth map to have the following properties: globally, landmark pixels associated with LiDAR points should possess the exact depths; locally, object shapes captured by neighboring 3D points, back-projected from the input depth map (cf. \autoref{eq_3D}), should be preserved.
\autoref{fig:PL++} (bottom) illustrates the algorithm.

\noindent\textbf{Input Matching.} We take as input the two point clouds from LiDAR (L) and Pseudo-LiDAR (PL) by stereo depth estimation. The latter is obtained by converting pixels $(u,v)$ with depth $z$ to 3D points $(x_u,y_v,z)$.
First, we characterize the local shapes by the directed K-nearest-neighbor (KNN) graph in the PL point cloud (using accelerated KD-Trees~\citep{shevtsov2007highly}) that connects each 3D point to its KNNs with appropriate weights.
Similarly, we can project the 3D LiDAR points onto pixel locations $(u,v)$ and match them to corresponding 3D stereo points. Without loss of generality, we assume that we are given ``ground truth'' LiDAR depth for the first $n$ points and no ground truth for the remaining $m$ points. We refer to the 3D stereo depth estimates as $Z\in\R^{n+m}$ and the LiDAR depth ground-truth as $G\in\R^n$.

\noindent\textbf{Edge weights.} To construct the KNN graph in 3D we ignore the LiDAR information on the first $n$ points and only use their predicted stereo depth in $Z$.
Let $\mathcal{N}_i$ denote the set of $k$ neighbors of the $i^{th}$ point.
Further, let $W\in{\R}^{(n+m)\times(n+m)}$ denote the weight matrix, where $W_{ij}$ denotes the edge-weight between points $i$ and $j$. Inspired by prior work in manifold learning~\citep{roweis2000nonlinear,WeinbergerPS05} we choose the weights to be the coefficients that reconstruct the depth of any point from the depths of its neighbors in $\mathcal{N}_i$. We can solve for these weights with the following constrained quadratic optimization problem:
\begin{align}
W=\argmin_{W} \|Z-WZ\|_2^2, \hspace{20pt}
\text{\ s.t.\ }  W\mathbf{1}=\mathbf{1} \textrm{ \ and \ } W_{ij}=0 \textrm{ if } j\notin {\mathcal N}_{i}.\label{eq:qp1}
\end{align}
Here $\mathbf{1}\in \R^{n+m}$ denotes the all-ones vector.
As long as we pick $k>3$ and the points are in general position there are infinitely many solutions that satisfy $Z=WZ$, and we pick the solution with the minimum $L_2$ norm (obtained with slight $L_2$ regularization).

\noindent\textbf{Depth Correction.} Let us denote the corrected depth values as $Z'\in \R^{n+m}$, with $Z'=[Z'_{L};Z'_{PL}]$ and $Z'_L\in \R^n$ and $Z'_{PL}\in\R^m$, where $Z'_L$ are the depth values of points with LiDAR ground-truth and $Z'_{PL}$ otherwise. For the $n$ points with LiDAR measurements we update the depth to the (ground truth) values $Z'_L=G$. We then solve for $Z'_{PL}$ given $G$ and the weighted KNN graph encoded in $W$.
Concretely, we update the remaining depths $Z'_{PL}$ such that the depth of any point $i$ can still be be reconstructed with high fidelity as a weighted sum of its KNNs' depths using the learned weights $W$; \ie if point $i:1\leq i\leq n$ is moved to its new depth $G_i$, then its neighbors in $\mathcal{N}_i$ must also be corrected such that $G_i\approx\sum_{j\in{\mathcal{N}}_i} W_{ij}Z'_j.$
Further, the neighbors' neighbors must be corrected and the depth of the few $n$ points propagates across the entire graph.
We can solve for the final $Z'$ directly with another quadratic optimization:
\begin{align}
Z'=\argmin_{Z'} \|Z' - WZ'\|^2,
\hspace{20pt} \text{\ s.t.\ }  Z'_{1:n}=G. \label{eq:qp2}
\end{align}
To illustrate the correction process, imagine the simplest case where the depth of only a single point ($n=1$) is updated  to $G_1=Z_1+\delta$. A new optimal depth for~\autoref{eq:qp2} is to move all the remaining points similarly, i.e.  $Z'=Z+\mathbf{1}\delta$:
as $Z=WZ$ and $W\mathbf{1}=\mathbf{1}$ we must have $W(Z+\mathbf{1}\delta)=Z+\mathbf{1}\delta$.
In the setting with $n>1$, the least-squares loss ensures a soft diffusion between the different LiDAR depth estimates. Both optimization problems in \autoref{eq:qp1} and \autoref{eq:qp2} can be solved exactly and efficiently with sparse matrix solvers. We summarize the procedure as an algorithm in the appendix.

From the view of graph-based manifold learning, our GDC algorithm is reminiscent of locally linear embeddings~\citep{roweis2000nonlinear} with landmarks to guide the final solution \citep{WeinbergerPS05}. \autoref{fig:redpurpleblue} illustrates vividly how the initial 3D point cloud from \SDN{} (purple) of a car in the KITTI dataset is corrected with a few sparse LiDAR measurements (yellow). The resulting points (blue) are right inside the ground-truth box and clearly show the contour of the car.
 \autoref{fig:depth_comp} shows the additional improvement from the \GDC{} (blue) over the pure \SDN{} depth estimates (see \autoref{sec::supp_stereo_error} for detailed numbers). The error (calculated only on non-landmark pixels) is corrected over the entire image where many regions have no LiDAR measurements. This is because that the pseudo-LiDAR point cloud is sufficiently dense and we choose $k$ to be large enough (in practice, we use $k=10$) such that the KNN graph is typically connected (or consists of few large connected components). See \autoref{sec::supp_connected_components} for more analysis.
 For objects such as cars the improvements through \GDC{} are far more pronounced, as these typically are touched by the four LiDAR beams and can be corrected effectively.

 
\section{Experiments}
 
\label{sec:exp}
\subsection{Setup}
\vspace{-1.5pt}
\label{sec:exp_setup}
\begin{table}[t]
	\centering
	\caption{\small \textbf{3D object detection results on KITTI validation.} We report \APBEV ~/ \AP (in \%) of the \textbf{car} category, corresponding to average precision of the bird's-eye view and 3D object detection. We arrange methods according to the input signals: M for monocular images, S for stereo images, L for 64-beam LiDAR, and L\# for \emph{sparse 4-beam} LiDAR. PL stands for \PL. \emph{Our \PL++ (PL++) with enhanced depth estimation --- \SDN and \GDC --- are in {\color{blue} blue}.} Methods with 64-beam LiDAR are in {\color{gray} gray}. Best viewed in color.} \label{tbMain}
	\tabcolsep 3pt
	\small
	\begin{tabular}{=l|+c|+c|+c|+c|+c|+c|+c}
		&  & \multicolumn{3}{c|}{IoU = 0.5} & \multicolumn{3}{c}{IoU = 0.7} \\ \cline{3-8}
		\multicolumn{1}{c|}{Detection algo} & Input & Easy & Moderate & Hard & Easy & Moderate & Hard \\ \hline
		\DOP & S & 55.0 / 46.0 & 41.3 / 34.6 & 34.6 / 30.1 & 12.6 / 6.6 \hspace{3pt} & 9.5 / 5.1 & 7.6 / 4.1 \\
		\MLFstereo & S & - & 53.7 / 47.4 & - & - & 19.5 / 9.8 \hspace{3pt} & - \\
		\SRCNN & S & 87.1 / 85.8 & 74.1 / 66.3  & 58.9  / 57.2 & 68.5 / 54.1 & 48.3 / 36.7 &  41.5 / 31.1 \\
		PL: \AVOD & S & 89.0 / 88.5 & 77.5 / 76.4 & 68.7 / 61.2&  74.9 / 61.9 &  56.8 / 45.3 & 49.0 / 39.0 \\
		PL: \vPIXOR & S & 89.0 / - \hspace{10pt} & 75.2 / - \hspace{10pt} & 67.3 / - \hspace{10pt} & 73.9 / - \hspace{10pt} & 54.0 / - \hspace{10pt} & 46.9 / - \hspace{10pt} \\
		PL: \PRCNN & S & 88.4 / 88.0 & 76.6 / 73.7 & 69.0 / 67.8 &  73.4 / 62.3 & 56.0 / 44.9 & 52.7 / 41.6 \\ \hline
		\rowstyle{\color{blue}}
		\hspace{-2.8pt}PL++: \AVOD & S & 89.4 / 89.0 & 79.0 / 77.8 & 70.1 / 69.1 & 77.0 / 63.2 & 63.7 / 46.8 & 56.0 / 39.8 \\
		\rowstyle{\color{blue}}
		\hspace{-2.8pt}PL++: \vPIXOR  & S & \textbf{89.9} / - \hspace{10pt} & 78.4 / - \hspace{10pt} & 74.7 / - \hspace{10pt} & 79.7 / - \hspace{10pt}   & 61.1 / - \hspace{10pt}  & 54.5 / - \hspace{10pt} \\
		\rowstyle{\color{blue}}
		\hspace{-2.8pt}PL++: \PRCNN  & S & 89.8 / \textbf{89.7} & \textbf{83.8} / \textbf{78.6} & \textbf{77.5} / \textbf{75.1} & \textbf{82.0} / \textbf{67.9} & \textbf{64.0} / \textbf{50.1} & \textbf{57.3} / \textbf{45.3} \\\hline
		\rowstyle{\color{blue}}
		\hspace{-2.8pt}PL++: \AVOD  & L\# + S & 90.2 / 90.1 & \textbf{87.7} / \textbf{86.9} & 79.8 / 79.2 & 86.8 / 70.7 & 76.6 / 56.2 & 68.7 / 53.4 \\
		\rowstyle{\color{blue}}
		\hspace{-2.8pt}PL++: \vPIXOR  & L\# + S & \textbf{95.1} / - \hspace{10pt} & 85.1 / - \hspace{10pt} & 78.3 / - \hspace{10pt} & 84.0 / - \hspace{10pt} & 71.0 / - \hspace{10pt} & 65.2 / - \hspace{10pt} \\
		\rowstyle{\color{blue}}
		\hspace{-2.8pt}PL++: \PRCNN & L\# + S & 90.3 / 90.3 & \textbf{87.7} / \textbf{86.9} & \textbf{84.6} / \textbf{84.2} & \textbf{88.2} / \textbf{75.1} & \textbf{76.9} / \textbf{63.8} & \textbf{73.4} / \textbf{57.4} \\
		\hline
		\rowstyle{\color{gray}}
		\hspace{-2.8pt}\AVOD & L + M & 90.5 / 90.5 & 89.4 / 89.2 & 88.5 / 88.2 & 89.4 / 82.8 & 86.5 / 73.5 & 79.3 / 67.1 \\
		\rowstyle{\color{gray}}
		\hspace{-2.8pt}\vPIXOR & L + M & 94.2 / - \hspace{10pt} & 86.7 / - \hspace{10pt} & 86.1 / - \hspace{10pt} & 85.2 / - \hspace{10pt} & 81.2 / - \hspace{10pt} & 76.1 / - \hspace{10pt} \\
		\rowstyle{\color{gray}}
		\hspace{-2.8pt}\PRCNN & L  & 97.3 / 97.3 & 89.9 / 89.8 & 89.4 / 89.3 &  90.2 / 89.2 & 87.9 / 78.9 & 85.5 / 77.9 \\
		\hline
	\end{tabular}
	\vspace{-10pt}
\end{table}

We refer to our combined method (\SDN{} and \GDC{}) for 3D object detection as \PL{}++ (PL++ in short). To analyze the contribution of each component, we evaluate \SDN{} and \GDC{} independently and jointly across several settings. For \GDC we set $k=10$ and consider adding signal from a (simulated) 4-beam LiDAR, unless stated otherwise.

\noindent\textbf{Dataset, Metrics, and Baselines.}
We evaluate on the KITTI dataset~\citep{geiger2013vision,geiger2012we}, which contains 7,481 and 7,518 images for training and testing. We follow~\citep{chen20153d} to separate the 7,481 images into 3,712 for training and 3,769 validation. For each (left) image, KITTI provides the corresponding right image, the 64-beam Velodyne LiDAR point cloud, the camera calibration matrices, and the bounding boxes.
We focus on 3D object detection and bird's-eye-view (BEV) localization and report results on the \emph{validation set}. Specifically, we focus on the ``car'' category, following~\cite{chen2017multi} and~\cite{xu2018pointfusion}. We report average precision (AP) with IoU (Intersection over Union) thresholds at 0.5 and 0.7. We denote AP for the 3D and BEV tasks by \AP and \APBEV.
KITTI defines the easy, moderate, and hard settings, in which objects with 2D box heights smaller than or occlusion/truncation levels larger than certain thresholds are disregarded.
We compare to four stereo-based detectors: \PL~(PL in short)~\citep{pseudoLiDAR},  \DOP~\citep{chen20153d}, \SRCNN\citep{li2019stereo}, and \MLFstereo~\citep{xu2018multi}.

\noindent\textbf{Stereo depth network (\SDN).}
We use \PSMNet~\citep{chang2018pyramid} as the backbone for our stereo depth estimation network (\SDN). We follow~\cite{pseudoLiDAR} to pre-train \SDN{} on the synthetic Scene Flow dataset~\citep{mayer2016large}
and fine-tune it on the 3,712 training images of KITTI. We obtain the depth ground truth by projecting the corresponding LiDAR points onto images. We also train a \PSMNet in the same way for comparison, which minimizes disparity error.

\noindent\textbf{3D object detection.} We apply three algorithms: \AVOD~\citep{ku2018joint}, \PIXOR~\citep{yang2018pixor}, and \PRCNN~\citep{shi2019pointrcnn}. All utilize information from LiDAR and/or monocular images. We use the released implementations of \AVOD ( specifically, \AVOD-FPN) and \PRCNN. We implement \PIXOR ourselves with a slight modification to include visual information (denoted as \vPIXOR). We train all models on the 3,712 training data from scratch by replacing the LiDAR points with pseudo-LiDAR data generated from stereo depth estimation. See the appendix for details.

\noindent\textbf{Sparser LiDAR.}
We simulate sparser LiDAR signal with fewer beams by first projecting the 64-beam LiDAR points onto a 2D plane of horizontal and vertical angles. We quantize the vertical angles into 64 levels with an interval of $0.4^{\circ}$, which is close to the SPEC of the 64-beam LiDAR. We keep points fallen into a subset of beams to mimic the sparser signal. See the appendix for details.


\subsection{Experimental results}
\label{sec:exp_result}

\noindent\textbf{Results on the KITTI val set.} We summarize the main results on KITTI object detection in \autoref{tbMain}.
Several important trends can be observed:
\textbf{1)} Our PL++ with enhanced depth estimations by \SDN and \GDC yields consistent improvement over PL across all settings; \textbf{2)} PL++ with \GDC refinement of 4-beam LiDAR (Input: L\# + S) performs significantly better than PL++ with only stereo inputs (Input: S); \textbf{3)} PL experiences a substantial drop in accuracy from IoU at 0.5 to 0.7 for the \emph{hard} setting. This suggests that while PL detects faraway objects, it mislocalizes them, likely placing them at the wrong depth. This causes the object to be considered a missed detection at higher overlap thresholds.
 Interestingly, here is where we experience the largest gain --- from PL: \PRCNN (\APBEV$=52.7$) to PL++: \PRCNN (\APBEV$=73.4$) with input as L\# + S. Note that the majority of the gain comes from \GDC{}, as PL++ with the stereo-only version only improving the score to $57.3$ \APBEV.
\textbf{4)} The gap between PL++ and LiDAR is at most $13\%$ \APBEV, even at the hard setting under IoU at 0.7.
\textbf{5)} For IoU at 0.5, with the aid of only 4 LiDAR beams, PL++ (SDN + GDC) achieves results comparable to models with 64-beam LiDAR signals.

\noindent\textbf{Results on the KITTI test set.}
\autoref{tbTest} summarizes results on the car category on the KITTI test set. We see a similar gap between our methods and LiDAR as on the validation set, suggesting that our improvement is not particular to the validation data. Our approach without LiDAR refinement (pure \SDN{}) is placed at the top position among all the image-based algorithms on the KITTI leaderboard.

\begin{table}

	\begin{minipage}[t]{.48\textwidth}
		\tabcolsep 2pt
		\fontsize{8}{9}\selectfont
		\centering
		\caption{\small \textbf{Results on the \textbf{car} category on the \emph{test} set.} We compare PL++ ({\color{blue}blue}) and 64-beam LiDAR ({\color{gray} gray}), using \PRCNN, and report \APBEV~/ \AP at IoU=0.7.}
		\vskip -5pt
		\begin{tabular}{l|c|c|c}
			\rowstyle{\color{black}} Input signal & Easy & Moderate & Hard \\ \hline
			\rowstyle{\color{blue}}
			PL++ (\SDN)  &  75.5 / 60.4 &  57.2 / 44.6 & 53.4 / 38.5 \\
			\rowstyle{\color{blue}}
			PL++ (\SDN+ \GDC)  &  83.8 / 68.5 &  73.5 / 54.7 & 66.5 / 51.2\\
			\rowstyle{\color{gray}}
			LiDAR  &  89.5 / 85.9 & 85.7 / 75.8 & 79.1 / 68.3\\
			\hline
		\end{tabular}
		\label{tbTest}
	\end{minipage}
	\hfill
	\begin{minipage}[t]{.48\textwidth}
		\tabcolsep 4pt
		\fontsize{8}{9}\selectfont
		\centering
		\caption{\small\textbf{Ablation study on depth estimation.} We report \APBEV ~/ \AP (in \%) of the \textbf{car} category at IoU$=0.7$ on KITTI validation. DL: depth loss.
		}
		\vskip -5pt
		\begin{tabular}{l|c|c|c}
			Stereo depth & Easy & Moderate & Hard \\ \hline
			\PSMNetpd                                                                  & 73.4 / 62.3                     & 56.0 / 44.9                     & 52.7 / 41.6    \\
			\PSMNetpd + DL                                                             & 80.1 / 65.5                     & 61.9 / 46.8                     & 56.0 / 43.0   \\
			\SDN    & \textbf{82.0} / \textbf{67.9} & \textbf{64.0} / \textbf{50.1} & \textbf{57.3} / \textbf{45.3}     \\ \hline
		\end{tabular}
		\label{tbl:ablation_sdn}
	\end{minipage}
\end{table}

\begin{table}
	\vspace{-5pt}
	\begin{minipage}{.48\textwidth}
		\tabcolsep 4pt
		\fontsize{8}{9}\selectfont
		\centering
		\caption{\small \textbf{Ablation study on leveraging sparse LiDAR.} We report \APBEV ~/ \AP (in \%) of the \textbf{car} category at IoU$=0.7$ on KITTI validation. L\#: 4-beam LiDAR signal alone. \SDN + L\#: pseudo-LiDAR with depths of landmark pixels replaced by 4-beam LiDAR. The best result of each column is in bold font.}  
		\label{tbl:ablation_lidar}
		\vskip-5pt
		\begin{tabular}{l|c|c|c}
		Stereo depth & Easy & Moderate & Hard \\ \hline
		\SDN    & 82.0 / 67.9 & 64.0 / 50.1 & 57.3 / 45.3   \\
		L\#    & 73.2 / 56.1 & 71.3 / 53.1 & 70.5 / 51.5   \\
		\SDN + L\#   & 86.3 / 72.0 & 73.0 / 56.1 & 67.4 / 54.1   \\
		\SDN + \GDC    & \textbf{88.2} / \textbf{75.1} & \textbf{76.9} / \textbf{63.8} & \textbf{73.4} / \textbf{57.4}  \\ \hline
	\end{tabular}
	\end{minipage}
	\hfill
	\begin{minipage}{.48\textwidth}
		\tabcolsep 4pt
		\fontsize{8}{9}\selectfont
		\centering
		\caption{\small \textbf{Results of \textbf{pedestrians} (top) and \textbf{cyclists} (bottom) on KITTI validation.} We apply \Frustum~\cite{qi2018frustum} and report \APBEV ~/ \AP (in \%) at IoU$=0.5$, following \cite{pseudoLiDAR}.}
		\label{tbPedestrian}
		\vskip -5pt
		\begin{tabular}{=l|+c|+c|+c}
		Stereo depth & Easy & Moderate & Hard \\ \hline
		\PSMNetpd & 41.3 / 33.8 & 34.9 / 27.4 & 30.1 / 24.0 \\
		\SDN & 48.7 / 40.9 & 40.4 / 32.9 & 34.9 / 28.8\\
		\SDN+ \GDC & \textbf{63.7} /  \textbf{53.6} &  \textbf{53.8} / \textbf{44.4} & \textbf{46.8} / \textbf{38.1} \\
		\hline
		\PSMNetpd & 47.6 / 41.3 & 29.9 / 25.2 & 27.0 / 24.9 \\
		\SDN & 49.3 / 44.6 &  30.4 / 28.7 & 28.6 / 26.4 \\
		\SDN+ \GDC & \textbf{65.7} / \textbf{60.8} & \textbf{45.8} / \textbf{40.8} & \textbf{42.8} / \textbf{38.0} \\
		\hline
	\end{tabular}
	\end{minipage}
	\vspace{-12pt}
\end{table}

In the following, we conduct a series of experiments to analyze the performance gain by our approaches and discuss several key observations. \emph{We mainly experiment with \PRCNN: we find that the results with \AVOD and \vPIXOR follow similar trends and thus include them in the appendix.}

\noindent\textbf{Depth loss and depth cost volume.} To turn a disparity network (\eg, \PSMNet) into \SDN, there are two changes: \textbf{1)} change the disparity loss into the depth loss; \textbf{2)} change the disparity cost volume into the depth cost volume. In \autoref{tbl:ablation_sdn}, we uncover the effect of these two changes separately. On the \APBEV/\AP (moderate) metric, the depth loss gives us a $6\% / 2\%$ improvement and the depth cost volume brings another $2\sim3\%$ gain\footnote{We note that, the degree of improvement brought by the depth loss and depth cost volume depends on the 3D detector in use. \autoref{tbl:ablation_sdn} suggests that the depth loss provides more gain than the depth cost volume (for \PRCNN). In \autoref{stbl:ablation_sdn}, we, however, see that the depth cost volume provides comparable or even bigger gain than the depth loss (for \vPIXOR and \AVOD). Nevertheless, \autoref{tbl:ablation_sdn} and \autoref{stbl:ablation_sdn} both suggest the compatibility of the two approaches: combining them leads to the best performance.}.

\noindent\textbf{Impact of sparse LiDAR beams.} We leverage 4-beam LiDAR to correct stereo depth using \GDC. However, it is possible that gains in 3D object detection come entirely from the new LiDAR sensor and that the stereo estimates are immaterial.
In \autoref{tbl:ablation_lidar}, we study this question by comparing the detection results against those of models using \textbf{1)} sole 4-beam LiDAR point clouds and
\textbf{2)} pseudo-LiDAR point clouds with depths of landmark pixels replaced by 4-beam LiDAR: \ie, in depth correction, we only correct depths of the landmark pixels without propagation.
It can be seen that 4-beam LiDAR itself performs fairly well on locating faraway objects but cannot capture nearby objects precisely, while simply replacing pseudo-LiDAR with LiDAR at the landmark pixels prevents the model from detecting faraway object accurately. In contrast, our proposed \GDC method effectively combines the merits of the two signals, achieving superior performance than using them alone.

\noindent\textbf{Pedestrian and cyclist detection.} For a fair comparison to~\citep{pseudoLiDAR}, we apply \Frustum \citep{qi2018frustum} for detecting pedestrians and cyclists. \autoref{tbPedestrian} shows the results: our methods significantly boosts the performance.

\begin{figure*}[htb!]
	\vspace{-10pt}
	\centering
	\includegraphics[width=.85\linewidth]{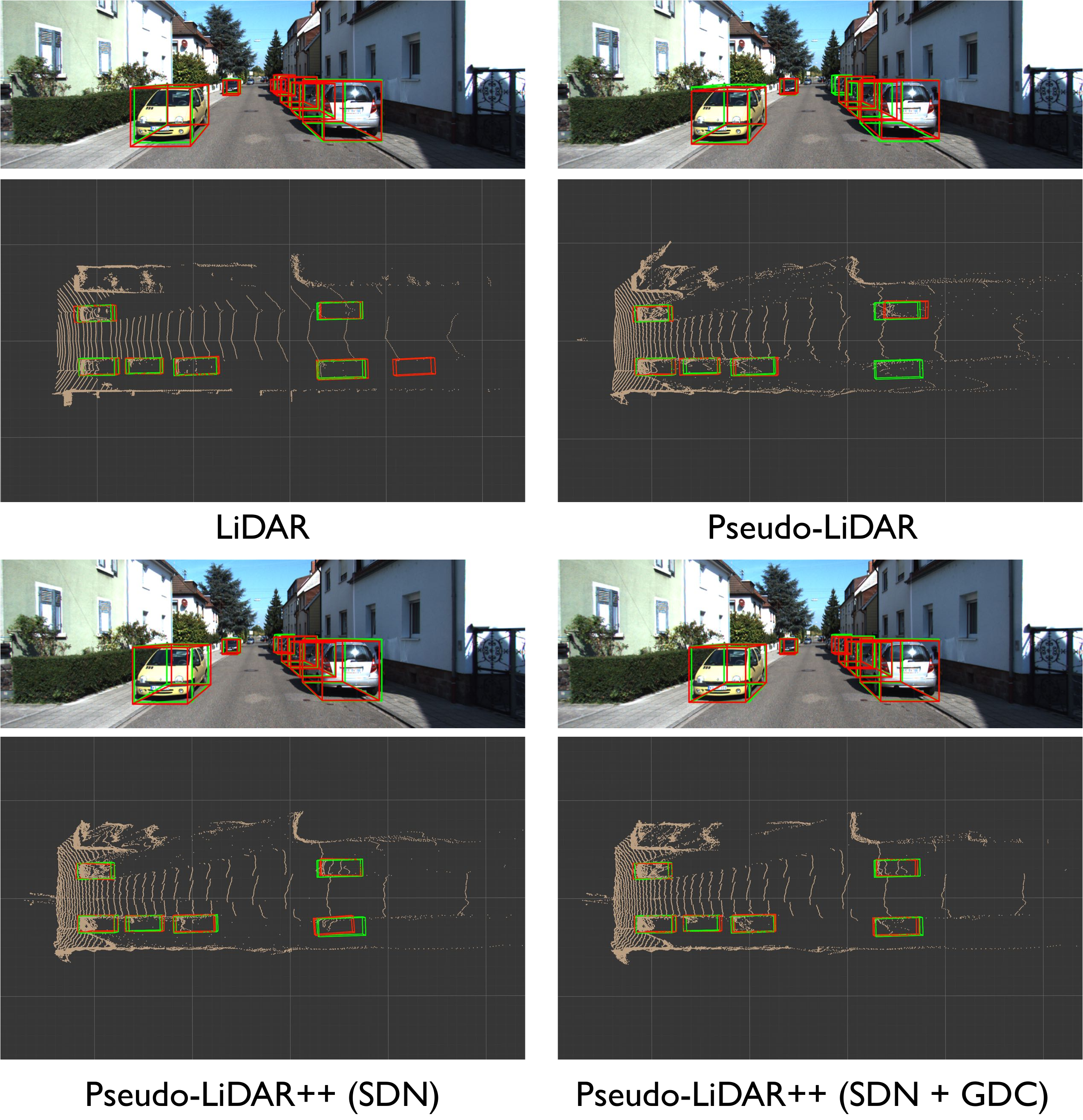}
	\caption{\textbf{Qualitative Comparison.} We show the detection results on a KITTI validation scene by \PRCNN with different input point clouds. We visualize them from both frontal-view images and bird's-eye view (BEV) point maps. Ground-truth boxes are in green and predicted bounding boxes are in red. The observer is at the left-hand side of the BEV map looking to the right. In other words, ground truth boxes on the right are more faraway (\ie, deeper) from the observer, and hence hard to localize. Best viewed in color. \label{fig:qualitative}}
	\vspace{-10pt}
\end{figure*}

\noindent\textbf{Qualitative visualization.}
In \autoref{fig:qualitative}, we show an qualitative comparison of detection results on a randomly chosen scene in the KITTI object validation set,
using \PRCNN (with confidence $> 0.95$) with different input signals. Specifically, we show the results from the frontal-view images and the bird's-eye view (BEV) point clouds. In the BEV map, the observer is on the left-hand side looking to the right. It can be seen that the point clouds generated by \PL++ (\SDN alone or \SDN+\GDC) align better with LiDAR than that generated by \PL (\PSMNet).
For nearby objects (\ie, bounding boxes close to the left in the BEV map), we see that \PRCNN with any point cloud performs fairly well in localization. However, for faraway objects (\ie, bounding boxes close to the right), \PL with depth estimated from \PSMNet predicts objects (red boxes) that are deviated from the ground truths (green boxes). Moreover, the noisy \PSMNet points also leads to false negatives. In contrast, the detected boxes by our \PL++, either with \SDN alone or with \SDN+\GDC, align pretty well with the ground truth boxes, justifying our targeted improvement in estimating faraway depths.

\noindent\textbf{Additional results, analyses, qualitative visualization and discussions.}
We provide results of \PL++ with fewer LiDAR beams, comparisons to depth completion methods, analysis on depth quality and detection accuracy, run time, failure cases, and more qualitative results in the appendix. With simple optimizations, \GDC runs in $90$ ms/frame using a single GPU ($7.7$ ms for KD-tree construction and search).

\vspace{-1pt}
\section{Conclusion}
\vspace{-1pt}
\label{sec:disc}
In this paper we made two contributions to improve the 3D object detection in autonomous vehicles without expensive LiDAR.  
First, we identify the disparity estimation as a main source of error for stereo-based systems and propose a novel approach to learn depth directly end-to-end instead of through disparity estimates. 
Second, we advocate that one should not use expensive LiDAR sensors to learn the local structure and depth of objects. Instead one can use commodity stereo cameras for the former and a cheap sparse LiDAR to correct the systematic bias in the resulting depth estimates. We provide a novel graph propagation algorithm that integrates the two data modalities and propagates the sparse yet accurate depth estimates using two sparse matrix solvers. 
The resulting system, \PL++ (\SDN + \GDC), performs almost on par with 64-beam LiDAR systems for \$75,000 but only requires 4 beams and two commodity cameras, which could be obtained with a total cost of less than \$1,000.

\section*{Acknowledgments}
This research is supported by grants from the National Science Foundation NSF (III-1618134, III-1526012, IIS-1149882, IIS-1724282, and TRIPODS-1740822), the Office of Naval Research DOD (N00014-17-1-2175), the Bill and Melinda Gates Foundation, and the Cornell Center for Materials Research with funding from the NSF MRSEC program (DMR-1719875). We are thankful for generous support by Zillow and SAP America Inc. We thank Gao Huang for helpful discussion.

\bibliographystyle{iclr2020_conference}
\bibliography{main}

\begin{thebibliography}{54}
\providecommand{\natexlab}[1]{#1}
\providecommand{\url}[1]{\texttt{#1}}
\expandafter\ifx\csname urlstyle\endcsname\relax
  \providecommand{\doi}[1]{doi: #1}\else
  \providecommand{\doi}{doi: \begingroup \urlstyle{rm}\Url}\fi

\bibitem[Chabot et~al.(2017)Chabot, Chaouch, Rabarisoa, Teuli{\`e}re, and
  Chateau]{chabot2017deep}
Florian Chabot, Mohamed Chaouch, Jaonary Rabarisoa, C{\'e}line Teuli{\`e}re,
  and Thierry Chateau.
\newblock Deep manta: A coarse-to-fine many-task network for joint 2d and 3d
  vehicle analysis from monocular image.
\newblock In \emph{CVPR}, 2017.

\bibitem[Chang \& Chen(2018)Chang and Chen]{chang2018pyramid}
Jia-Ren Chang and Yong-Sheng Chen.
\newblock Pyramid stereo matching network.
\newblock In \emph{CVPR}, 2018.

\bibitem[Chen et~al.(2015)Chen, Kundu, Zhu, Berneshawi, Ma, Fidler, and
  Urtasun]{chen20153d}
Xiaozhi Chen, Kaustav Kundu, Yukun Zhu, Andrew~G Berneshawi, Huimin Ma, Sanja
  Fidler, and Raquel Urtasun.
\newblock 3d object proposals for accurate object class detection.
\newblock In \emph{NIPS}, 2015.

\bibitem[Chen et~al.(2016)Chen, Kundu, Zhang, Ma, Fidler, and
  Urtasun]{chen2016monocular}
Xiaozhi Chen, Kaustav Kundu, Ziyu Zhang, Huimin Ma, Sanja Fidler, and Raquel
  Urtasun.
\newblock Monocular 3d object detection for autonomous driving.
\newblock In \emph{CVPR}, 2016.

\bibitem[Chen et~al.(2017)Chen, Ma, Wan, Li, and Xia]{chen2017multi}
Xiaozhi Chen, Huimin Ma, Ji~Wan, Bo~Li, and Tian Xia.
\newblock Multi-view 3d object detection network for autonomous driving.
\newblock In \emph{CVPR}, 2017.

\bibitem[Chen et~al.(2018)Chen, Kundu, Zhu, Ma, Fidler, and
  Urtasun]{chen20183d}
Xiaozhi Chen, Kaustav Kundu, Yukun Zhu, Huimin Ma, Sanja Fidler, and Raquel
  Urtasun.
\newblock 3d object proposals using stereo imagery for accurate object class
  detection.
\newblock \emph{IEEE TPAMI}, 40\penalty0 (5):\penalty0 1259--1272, 2018.

\bibitem[Cheng et~al.(2018)Cheng, Wang, and Yang]{cheng2018depth}
Xinjing Cheng, Peng Wang, and Ruigang Yang.
\newblock Depth estimation via affinity learned with convolutional spatial
  propagation network.
\newblock In \emph{ECCV}, 2018.

\bibitem[Du et~al.(2018)Du, Ang~Jr, Karaman, and Rus]{du2018general}
Xinxin Du, Marcelo~H Ang~Jr, Sertac Karaman, and Daniela Rus.
\newblock A general pipeline for 3d detection of vehicles.
\newblock In \emph{ICRA}, 2018.

\bibitem[Engelcke et~al.(2017)Engelcke, Rao, Wang, Tong, and
  Posner]{engelcke2017vote3deep}
Martin Engelcke, Dushyant Rao, Dominic~Zeng Wang, Chi~Hay Tong, and Ingmar
  Posner.
\newblock Vote3deep: Fast object detection in 3d point clouds using efficient
  convolutional neural networks.
\newblock In \emph{ICRA}, 2017.

\bibitem[Fu et~al.(2018)Fu, Gong, Wang, Batmanghelich, and Tao]{fu2018deep}
Huan Fu, Mingming Gong, Chaohui Wang, Kayhan Batmanghelich, and Dacheng Tao.
\newblock Deep ordinal regression network for monocular depth estimation.
\newblock In \emph{CVPR}, pp.\  2002--2011, 2018.

\bibitem[Geiger et~al.(2012)Geiger, Lenz, and Urtasun]{geiger2012we}
Andreas Geiger, Philip Lenz, and Raquel Urtasun.
\newblock Are we ready for autonomous driving? the kitti vision benchmark
  suite.
\newblock In \emph{CVPR}, 2012.

\bibitem[Geiger et~al.(2013)Geiger, Lenz, Stiller, and
  Urtasun]{geiger2013vision}
Andreas Geiger, Philip Lenz, Christoph Stiller, and Raquel Urtasun.
\newblock Vision meets robotics: The kitti dataset.
\newblock \emph{The International Journal of Robotics Research}, 32\penalty0
  (11):\penalty0 1231--1237, 2013.

\bibitem[Godard et~al.(2017)Godard, Mac~Aodha, and
  Brostow]{godard2017unsupervised}
Cl{\'e}ment Godard, Oisin Mac~Aodha, and Gabriel~J Brostow.
\newblock Unsupervised monocular depth estimation with left-right consistency.
\newblock In \emph{CVPR}, 2017.

\bibitem[He et~al.(2016)He, Zhang, Ren, and Sun]{he2016deep}
Kaiming He, Xiangyu Zhang, Shaoqing Ren, and Jian Sun.
\newblock Deep residual learning for image recognition.
\newblock In \emph{CVPR}, 2016.

\bibitem[He et~al.(2017)He, Gkioxari, Doll{\'a}r, and Girshick]{he2017mask}
Kaiming He, Georgia Gkioxari, Piotr Doll{\'a}r, and Ross Girshick.
\newblock Mask r-cnn.
\newblock In \emph{ICCV}, 2017.

\bibitem[Ku et~al.(2018)Ku, Mozifian, Lee, Harakeh, and Waslander]{ku2018joint}
Jason Ku, Melissa Mozifian, Jungwook Lee, Ali Harakeh, and Steven Waslander.
\newblock Joint 3d proposal generation and object detection from view
  aggregation.
\newblock In \emph{IROS}, 2018.

\bibitem[Lang et~al.(2019)Lang, Vora, Caesar, Zhou, Yang, and
  Beijbom]{lang2019pointpillars}
Alex~H Lang, Sourabh Vora, Holger Caesar, Lubing Zhou, Jiong Yang, and Oscar
  Beijbom.
\newblock Pointpillars: Fast encoders for object detection from point clouds.
\newblock In \emph{CVPR}, 2019.

\bibitem[Li(2017)]{li20173d}
Bo~Li.
\newblock 3d fully convolutional network for vehicle detection in point cloud.
\newblock In \emph{IROS}, 2017.

\bibitem[Li et~al.(2016)Li, Zhang, and Xia]{li2016vehicle}
Bo~Li, Tianlei Zhang, and Tian Xia.
\newblock Vehicle detection from 3d lidar using fully convolutional network.
\newblock In \emph{Robotics: Science and Systems}, 2016.

\bibitem[Li et~al.(2019{\natexlab{a}})Li, Ouyang, Sheng, Zeng, and
  Wang]{li2019gs3d}
Buyu Li, Wanli Ouyang, Lu~Sheng, Xingyu Zeng, and Xiaogang Wang.
\newblock Gs3d: An efficient 3d object detection framework for autonomous
  driving.
\newblock In \emph{CVPR}, 2019{\natexlab{a}}.

\bibitem[Li et~al.(2019{\natexlab{b}})Li, Chen, and Shen]{li2019stereo}
Peiliang Li, Xiaozhi Chen, and Shaojie Shen.
\newblock Stereo r-cnn based 3d object detection for autonomous driving.
\newblock In \emph{CVPR}, 2019{\natexlab{b}}.

\bibitem[Liang et~al.(2018)Liang, Yang, Wang, and Urtasun]{liang2018deep}
Ming Liang, Bin Yang, Shenlong Wang, and Raquel Urtasun.
\newblock Deep continuous fusion for multi-sensor 3d object detection.
\newblock In \emph{ECCV}, 2018.

\bibitem[Lin et~al.(2017)Lin, Doll{\'a}r, Girshick, He, Hariharan, and
  Belongie]{lin2017feature}
Tsung-Yi Lin, Piotr Doll{\'a}r, Ross~B Girshick, Kaiming He, Bharath Hariharan,
  and Serge~J Belongie.
\newblock Feature pyramid networks for object detection.
\newblock In \emph{CVPR}, volume~1, pp.\ ~4, 2017.

\bibitem[Ma et~al.(2019)Ma, Cavalheiro, and Karaman]{ma2019self}
Fangchang Ma, Guilherme~Venturelli Cavalheiro, and Sertac Karaman.
\newblock Self-supervised sparse-to-dense: self-supervised depth completion
  from lidar and monocular camera.
\newblock In \emph{ICRA}, 2019.

\bibitem[Mayer et~al.(2016)Mayer, Ilg, Hausser, Fischer, Cremers, Dosovitskiy,
  and Brox]{mayer2016large}
Nikolaus Mayer, Eddy Ilg, Philip Hausser, Philipp Fischer, Daniel Cremers,
  Alexey Dosovitskiy, and Thomas Brox.
\newblock A large dataset to train convolutional networks for disparity,
  optical flow, and scene flow estimation.
\newblock In \emph{CVPR}, 2016.

\bibitem[Meyer et~al.(2019{\natexlab{a}})Meyer, Charland, Hegde, Laddha, and
  Vallespi-Gonzalez]{meyer2019sensor}
Gregory~P Meyer, Jake Charland, Darshan Hegde, Ankit Laddha, and Carlos
  Vallespi-Gonzalez.
\newblock Sensor fusion for joint 3d object detection and semantic
  segmentation.
\newblock \emph{arXiv preprint arXiv:1904.11466}, 2019{\natexlab{a}}.

\bibitem[Meyer et~al.(2019{\natexlab{b}})Meyer, Laddha, Kee, Vallespi-Gonzalez,
  and Wellington]{meyer2019lasernet}
Gregory~P Meyer, Ankit Laddha, Eric Kee, Carlos Vallespi-Gonzalez, and Carl~K
  Wellington.
\newblock Lasernet: An efficient probabilistic 3d object detector for
  autonomous driving.
\newblock In \emph{CVPR}, 2019{\natexlab{b}}.

\bibitem[Mousavian et~al.(2017)Mousavian, Anguelov, Flynn, and
  Ko{\v{s}}eck{\'a}]{mousavian20173d}
Arsalan Mousavian, Dragomir Anguelov, John Flynn, and Jana Ko{\v{s}}eck{\'a}.
\newblock 3d bounding box estimation using deep learning and geometry.
\newblock In \emph{CVPR}, 2017.

\bibitem[Pham \& Jeon(2017)Pham and Jeon]{pham2017robust}
Cuong~Cao Pham and Jae~Wook Jeon.
\newblock Robust object proposals re-ranking for object detection in autonomous
  driving using convolutional neural networks.
\newblock \emph{Signal Processing: Image Communication}, 53:\penalty0 110--122,
  2017.

\bibitem[Qi et~al.(2017{\natexlab{a}})Qi, Su, Mo, and Guibas]{qi2017pointnet}
Charles~R Qi, Hao Su, Kaichun Mo, and Leonidas~J Guibas.
\newblock Pointnet: Deep learning on point sets for 3d classification and
  segmentation.
\newblock In \emph{CVPR}, 2017{\natexlab{a}}.

\bibitem[Qi et~al.(2018)Qi, Liu, Wu, Su, and Guibas]{qi2018frustum}
Charles~R Qi, Wei Liu, Chenxia Wu, Hao Su, and Leonidas~J Guibas.
\newblock Frustum pointnets for 3d object detection from rgb-d data.
\newblock In \emph{CVPR}, 2018.

\bibitem[Qi et~al.(2017{\natexlab{b}})Qi, Yi, Su, and Guibas]{qi2017pointnet++}
Charles~Ruizhongtai Qi, Li~Yi, Hao Su, and Leonidas~J Guibas.
\newblock Pointnet++: Deep hierarchical feature learning on point sets in a
  metric space.
\newblock In \emph{NIPS}, 2017{\natexlab{b}}.

\bibitem[Ren et~al.(2015)Ren, He, Girshick, and Sun]{ren2015faster}
Shaoqing Ren, Kaiming He, Ross Girshick, and Jian Sun.
\newblock Faster r-cnn: Towards real-time object detection with region proposal
  networks.
\newblock In \emph{NIPS}, 2015.

\bibitem[Roweis \& Saul(2000)Roweis and Saul]{roweis2000nonlinear}
Sam~T Roweis and Lawrence~K Saul.
\newblock Nonlinear dimensionality reduction by locally linear embedding.
\newblock \emph{science}, 2000.

\bibitem[Shevtsov et~al.(2007)Shevtsov, Soupikov, and
  Kapustin]{shevtsov2007highly}
Maxim Shevtsov, Alexei Soupikov, and Alexander Kapustin.
\newblock Highly parallel fast kd-tree construction for interactive ray tracing
  of dynamic scenes.
\newblock In \emph{Computer Graphics Forum}, volume~26, pp.\  395--404. Wiley
  Online Library, 2007.

\bibitem[Shi et~al.(2019)Shi, Wang, and Li]{shi2019pointrcnn}
Shaoshuai Shi, Xiaogang Wang, and Hongsheng Li.
\newblock Pointrcnn: 3d object proposal generation and detection from point
  cloud.
\newblock In \emph{CVPR}, 2019.

\bibitem[Srivastava et~al.(2019)Srivastava, Jurie, and
  Sharma]{srivastava2019learning}
Siddharth Srivastava, Frederic Jurie, and Gaurav Sharma.
\newblock Learning 2d to 3d lifting for object detection in 3d for autonomous
  vehicles.
\newblock In \emph{IROS}, 2019.

\bibitem[Tomasello et~al.(2018)Tomasello, Sidhu, Shen, Moskewicz, Redmon,
  Joshi, Phadte, Jain, and Iandola]{tomasello2018dscnet}
Paden Tomasello, Sammy Sidhu, Anting Shen, Matthew~W Moskewicz, Nobie Redmon,
  Gataryi Joshi, Romi Phadte, Paras Jain, and Forrest Iandola.
\newblock Dscnet: Replicating lidar point clouds with deep sensor cloning.
\newblock \emph{arXiv preprint arXiv:1811.07070}, 2018.

\bibitem[Torres-Mendez \& Dudek(2004)Torres-Mendez and
  Dudek]{torres2004statistical}
Luz~Abril Torres-Mendez and Gregory Dudek.
\newblock Statistical inference and synthesis in the image domain for mobile
  robot environment modeling.
\newblock In \emph{IROS}, 2004.

\bibitem[Wang et~al.(2018)Wang, Wang, Lin, Tsai, Chiu, and Sun]{wang2018plug}
Tsun-Hsuan Wang, Fu-En Wang, Juan-Ting Lin, Yi-Hsuan Tsai, Wei-Chen Chiu, and
  Min Sun.
\newblock Plug-and-play: Improve depth estimation via sparse data propagation.
\newblock \emph{arXiv preprint arXiv:1812.08350}, 2018.

\bibitem[Wang et~al.(2019{\natexlab{a}})Wang, Chao, Garg, Hariharan, Campbell,
  and Weinberger]{pseudoLiDAR}
Yan Wang, Wei-Lun Chao, Divyansh Garg, Bharath Hariharan, Mark Campbell, and
  Kilian~Q. Weinberger.
\newblock Pseudo-lidar from visual depth estimation: Bridging the gap in 3d
  object detection for autonomous driving.
\newblock In \emph{CVPR}, 2019{\natexlab{a}}.

\bibitem[Wang et~al.(2019{\natexlab{b}})Wang, Lai, Huang, Wang, van~der Maaten,
  Campbell, and Weinberger]{wang2018anytime}
Yan Wang, Zihang Lai, Gao Huang, Brian~H Wang, Laurens van~der Maaten, Mark
  Campbell, and Kilian~Q Weinberger.
\newblock Anytime stereo image depth estimation on mobile devices.
\newblock In \emph{ICRA}, 2019{\natexlab{b}}.

\bibitem[Weinberger et~al.(2005)Weinberger, Packer, and Saul]{WeinbergerPS05}
Kilian~Q. Weinberger, Benjamin Packer, and Lawrence~K. Saul.
\newblock Nonlinear dimensionality reduction by semidefinite programming and
  kernel matrix factorization.
\newblock In \emph{AISTATS}, 2005.

\bibitem[Xiang et~al.(2015)Xiang, Choi, Lin, and Savarese]{xiang2015data}
Yu~Xiang, Wongun Choi, Yuanqing Lin, and Silvio Savarese.
\newblock Data-driven 3d voxel patterns for object category recognition.
\newblock In \emph{CVPR}, 2015.

\bibitem[Xiang et~al.(2017)Xiang, Choi, Lin, and
  Savarese]{xiang2017subcategory}
Yu~Xiang, Wongun Choi, Yuanqing Lin, and Silvio Savarese.
\newblock Subcategory-aware convolutional neural networks for object proposals
  and detection.
\newblock In \emph{WACV}, 2017.

\bibitem[Xiaojin \& Zoubin(2002)Xiaojin and Zoubin]{xiaojin2002learning}
Zhu Xiaojin and Ghahramani Zoubin.
\newblock Learning from labeled and unlabeled data with label propagation.
\newblock \emph{Tech. Rep., Technical Report CMU-CALD-02--107, Carnegie Mellon
  University}, 2002.

\bibitem[Xu \& Chen(2018)Xu and Chen]{xu2018multi}
Bin Xu and Zhenzhong Chen.
\newblock Multi-level fusion based 3d object detection from monocular images.
\newblock In \emph{CVPR}, 2018.

\bibitem[Xu et~al.(2018)Xu, Anguelov, and Jain]{xu2018pointfusion}
Danfei Xu, Dragomir Anguelov, and Ashesh Jain.
\newblock Pointfusion: Deep sensor fusion for 3d bounding box estimation.
\newblock In \emph{CVPR}, 2018.

\bibitem[Yamaguchi et~al.(2014)Yamaguchi, McAllester, and
  Urtasun]{yamaguchi2014efficient}
Koichiro Yamaguchi, David McAllester, and Raquel Urtasun.
\newblock Efficient joint segmentation, occlusion labeling, stereo and flow
  estimation.
\newblock In \emph{ECCV}, 2014.

\bibitem[Yan et~al.(2018)Yan, Mao, and Li]{yan2018second}
Yan Yan, Yuxing Mao, and Bo~Li.
\newblock Second: Sparsely embedded convolutional detection.
\newblock \emph{Sensors}, 18\penalty0 (10):\penalty0 3337, 2018.

\bibitem[Yang et~al.(2018{\natexlab{a}})Yang, Liang, and
  Urtasun]{yang2018hdnet}
Bin Yang, Ming Liang, and Raquel Urtasun.
\newblock Hdnet: Exploiting hd maps for 3d object detection.
\newblock In \emph{Conference on Robot Learning}, pp.\  146--155,
  2018{\natexlab{a}}.

\bibitem[Yang et~al.(2018{\natexlab{b}})Yang, Luo, and Urtasun]{yang2018pixor}
Bin Yang, Wenjie Luo, and Raquel Urtasun.
\newblock Pixor: Real-time 3d object detection from point clouds.
\newblock In \emph{CVPR}, 2018{\natexlab{b}}.

\bibitem[Yang et~al.(2019)Yang, Wong, and Soatto]{yang2019dense}
Yanchao Yang, Alex Wong, and Stefano Soatto.
\newblock Dense depth posterior (ddp) from single image and sparse range.
\newblock In \emph{CVPR}, 2019.

\bibitem[Zhou \& Tuzel(2018)Zhou and Tuzel]{zhou2018voxelnet}
Yin Zhou and Oncel Tuzel.
\newblock Voxelnet: End-to-end learning for point cloud based 3d object
  detection.
\newblock In \emph{CVPR}, 2018.

\end{thebibliography}

\clearpage
\appendix
\begin{center}
	\textbf{\Large Appendix}
\end{center}
We provide details omitted in the main text.

\begin{itemize}
	\item \autoref{sec:sup_sdn}: details on constructing the depth cost volume (\autoref{sec:approach} of the main paper).
	\item \autoref{sec:sup_gdc}: detailed implementation of the \GDC algorithm (\autoref{sec:depth_correction} of the main paper).
	\item \autoref{ssec:ES}: additional details of experimental setups (\autoref{sec:exp_setup} of the main paper).
	\item \autoref{ssec:AR}: additional results, analyses, and discussions (\autoref{sec:exp_result} of the main paper).
\end{itemize}

\section{Depth Cost Volume}
\label{sec:sup_sdn}
With \autoref{eq_disp_depth}, we know where each grid $(u,v,z)$ in $C_\text{depth}$ corresponds to in $C_\text{disp}$ (may not be on a grid). We can then obtain features for each grid in $C_\text{depth}$ (\ie, $C_\text{depth}(u,v,z,:)$) by bilinear interpolation over features on grids of $C_\text{disp}$ around the non-grid location (\ie, $\left(u,v,\cfrac{f_U\times b}{z}\right)$). We applied the ``grid\_sample'' function in PyTorch for bilinear interpolation.

We use \PSMNet~\citep{chang2018pyramid} as the backbone for our stereo depth estimation network (\SDN). The only change is to construct the depth cost volume before performing 3D convolutions.

\section{Graph-based Depth Correction (\GDC) Algorithm}
\label{sec:sup_gdc}
Here we present the \GDC algorithm in detail (see \autoref{alg::GDC}). The two steps described in the main paper can be easily turned into two (sparse) linear systems and then solved by using Lagrange multipliers. For the first step (\ie, \autoref{eq:qp1}), we solve the same problem as in the main text but we switch the objective to minimizing the $L_2$-norm of $W$ and set $Z-WZ=0$ as a constraint\footnote{These two problems yield identical solutions but we found the second one is easier to solve in practice. We note that, \autoref{eq:qp1} is an under-constrained problem, with infinitely many solutions. To identify a single solution, we add a small $L_2$ regularization term to the objective (as mentioned in the main text). }. For the second step (\ie, \autoref{eq:qp2}), we use the \emph{Conjugate Gradient} (CG) to iteratively solve the sparse linear system.

\begin{algorithm*}
	\caption{Graph-based depth correction (\GDC). ``$;$'' stands for column-wise concatenation. \label{alg::GDC}}
	\KwIn{Stereo depth map $Z\in\R^{(n+m)\times 1}$, the corresponding pseudo-LiDAR (PL) point cloud $P \in \mathbb{R}^{(n+m)\times 3}$, and LiDAR depths $G\in \R^{n\times 1}$ on the first the $n$ pixels.}
	\KwOut{Corrected depth map $Z' \in \R^{(n+m)\times 1}$}
	\SetAlgoLined
	\SetKwBlock{Begin}{function}{end function}
	\Begin($\text{\GDC} {(} {Z}, P, G, K {)}$){
		Solve: $W = \arg\min_{W\in\R^{(n+m)\times(n+m)}} \|W\|^2$\\
		\hspace{2.55em} s.t. \hspace{0.7em} ${Z} - W\cdot {Z} = 0$,\\
		\hspace{5em} $W_{ij} = 0$ if $j\notin\mathcal{N}_i$ (\ie, the set of neighbors of the $i^{th}$ point) according to $P$,\\
		\hspace{5em} $\sum_{j}W_{ij} = 1$ for $\forall i = 1, \dots, n+m$.\\
		Solve: $Z_{PL}' = \arg\min_{Z_{PL}'\in \R^{m\times 1}} \|[G; Z_{PL}'] - W[G; Z_{PL}']\|^2$\\
		\Return $[G; Z_{PL}']$
	}
\end{algorithm*}

\section{Experimental Setup}
\label{ssec:ES}
\subsection{Sparse LiDAR generation}
\label{ssec:sparse}
\begin{figure*}[!htb]
	\centering
	\begin{subfigure}[b]{0.29\textwidth}
	    \includegraphics[width=\textwidth]{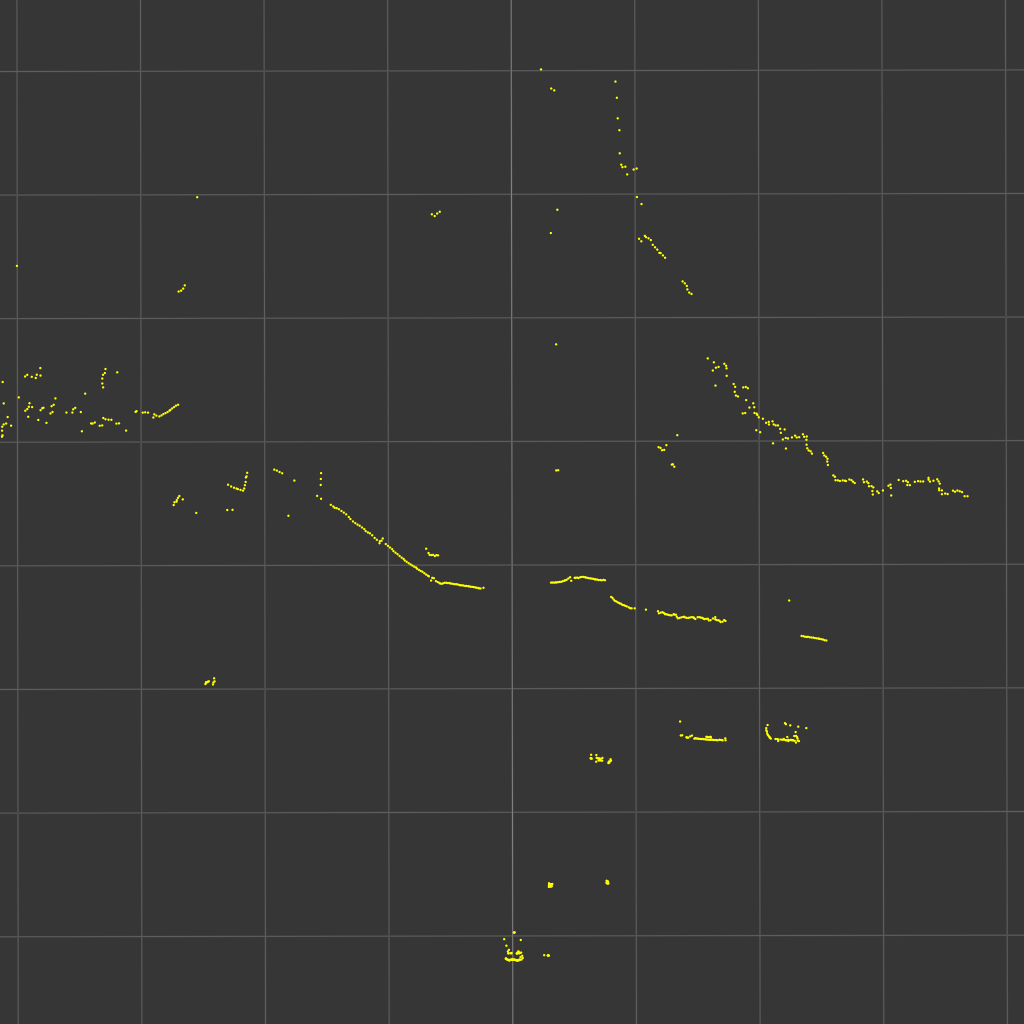}
	    \caption{2-beam}
	\end{subfigure}
	\begin{subfigure}[b]{0.29\textwidth}
	    \includegraphics[width=\textwidth]{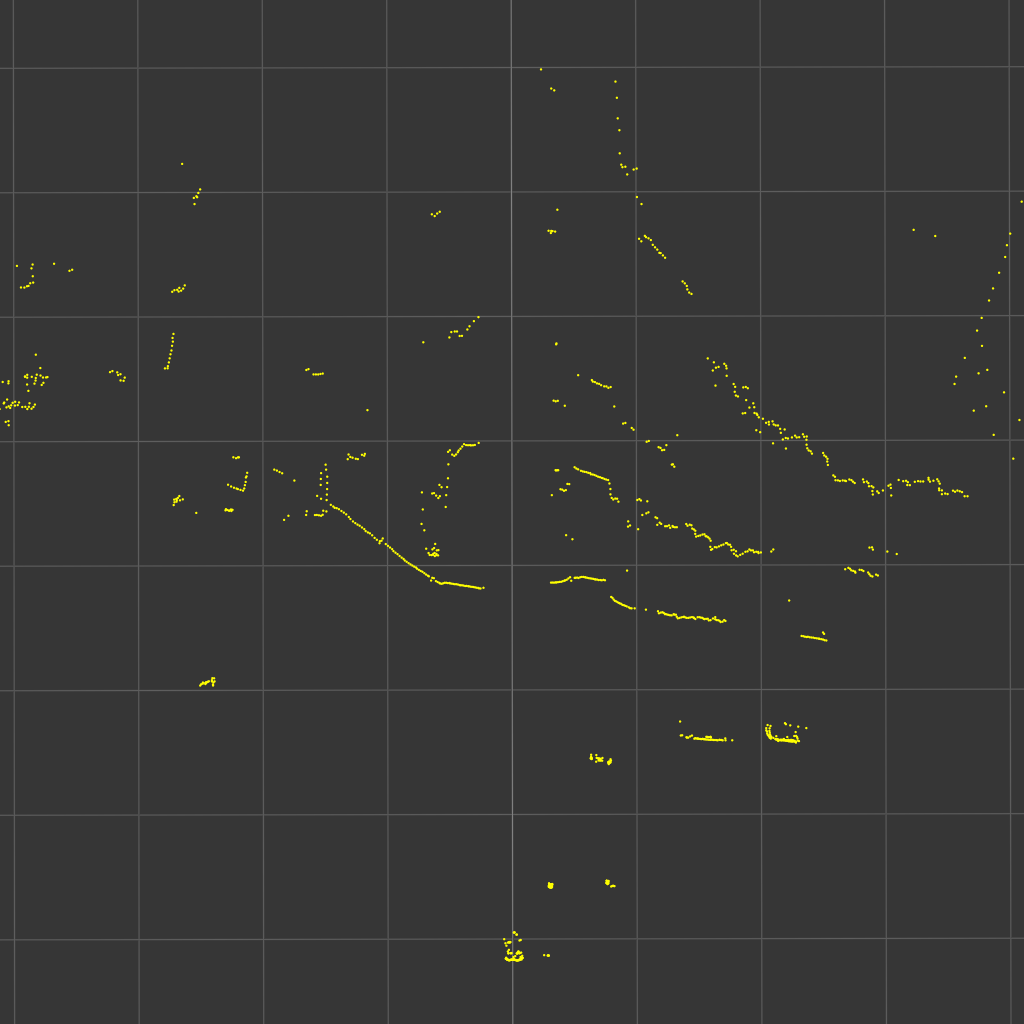}
	    \caption{4-beam}
	\end{subfigure}
	\begin{subfigure}[b]{0.29\textwidth}
	    \includegraphics[width=\textwidth]{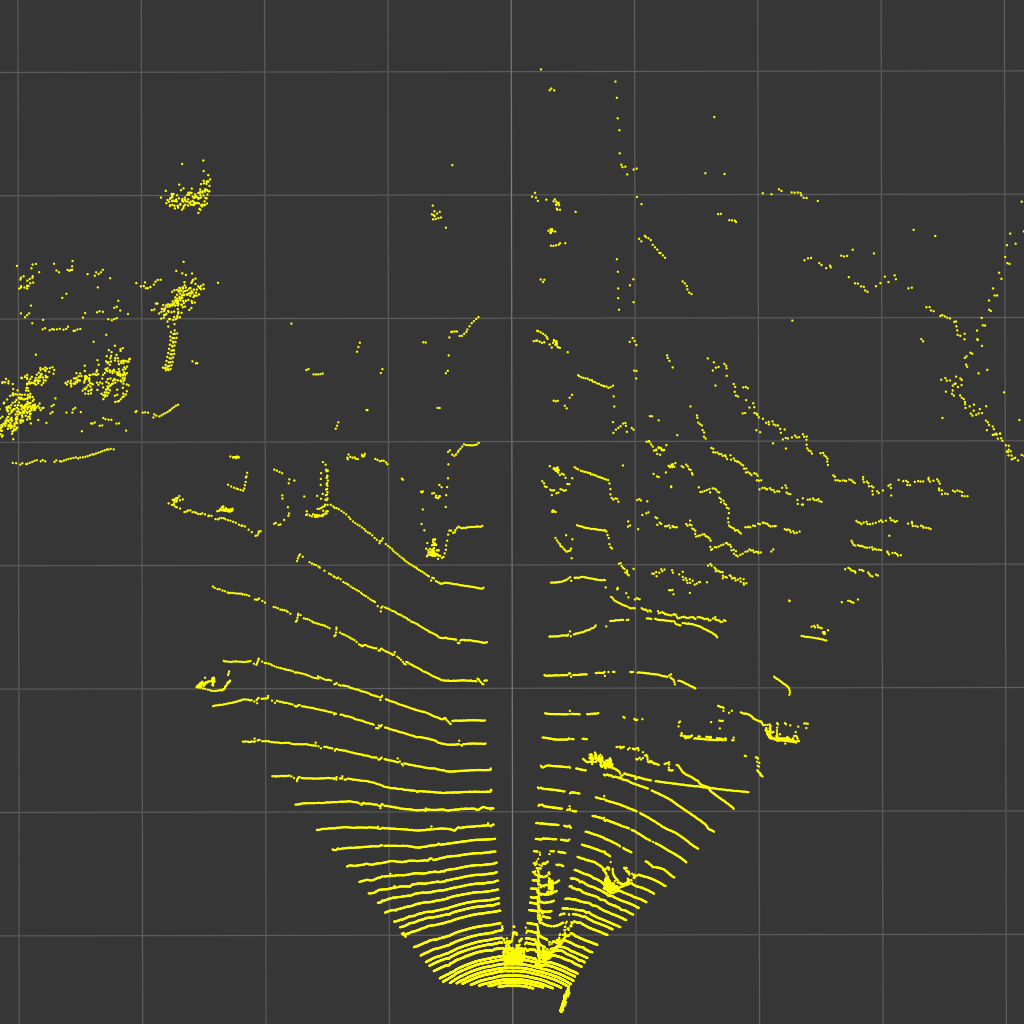}
	    \caption{64-beam (full)}
	\end{subfigure}
	\caption{\textbf{Bird's-eye views of sparsified LiDAR on an example scene.} The observer is on the bottom side looking up. We filter out points invisible from the left image. (One floor square is 10m $\times$ 10m.) \label{fig::bevsparsed}}
\end{figure*}

In this section, we explain how we generate sparser LiDAR with fewer beams from a 64-beam LiDAR point cloud from KITTI dataset in detail. For every point $(x_i,y_i,z_i)\in\R^3$ of the point cloud in one scene (in LiDAR coordinate system ($x$: front, $y$: left, $z$: up, and $(0, 0, 0)$ is the location of the LiDAR sensor)), we compute the elevation angle $\theta_i$ to the LiDAR sensor as
\begin{displaymath}
	\theta_i = \arg\cos\left(\frac{\sqrt{x_i^2 + y_i^2}}{\sqrt{x_i^2 + y_i^2 + z_i^2}}\right).
\end{displaymath}
We order the points by their elevation angles and slice them into separate lines by step $0.4^{\circ}$, starting from $-23.6^{\circ}$ (close to the Velodyne 64-beam LiDAR SPEC). We select LiDAR points whose elevation angles fall within $[-2.4^{\circ}, -2.0^{\circ}) \cup [-0.8^{\circ}, -0.4^{\circ})$ to be the 2-beam LiDAR signal, and similarly $ [-2.4^{\circ}, -2.0^{\circ}) \cup [-1.6^{\circ}, -1.2^{\circ}) \cup [-0.8^{\circ}, -0.4^{\circ}) \cup [0.0^{\circ}, 0.4^{\circ})$ to be the 4-beam LiDAR signal. We choose them in such a way that consecutive lines has a $0.8^{\circ}$ interval, following the SPEC of the ``cheap'' 4-beam LiDAR ScaLa.  We visualize these sparsified LiDAR point clouds from the bird's-eye view on one example scene in \autoref{fig::bevsparsed}.

\subsection{3D object detection algorithms}
\label{ssec:3D}
In this section, we provide more details about the way we train 3D object detection models on pseudo-LiDAR point clouds.
For \AVOD, we use the same model as in~\citep{pseudoLiDAR}.
For \PRCNN, we use the implementation provided by the authors. Since the \PRCNN model exploits the sparse nature of LiDAR point clouds, when training it with pseudo-LiDAR input, we will first sparsify the point clouds into 64 beams using the method described in \autoref{ssec:sparse}.
For \vPIXOR, we implement the same base model structure and data augmentation specified by \cite{yang2018pixor}, but without the ``decode fine-tune'' step and focal loss. Inspired by the trick in~\citep{liang2018deep}, we add another image feature (ResNet-18 by~\cite{he2016deep}) branch along the LiDAR branch, and concatenate the corresponding image features onto the LiDAR branch at each stage. We train \vPIXOR using RMSProp with momentum $0.9$, learning rate $10^{-5}$ (decay by 10 after 50 and 80 epochs) for 90 epochs. The BEV evaluation results are similar to the reported results (see \autoref{tbMain}).

\section{Additional Results, Analyses, and Discussions}
\label{ssec:AR}

\subsection{Ablation study}
\label{ssec:AS}

In \autoref{stbl:ablation_sdn} and \autoref{stbl:ablation_lidar} we provide more experimental results aligned with experiments in \autoref{sec:exp_result} of the main paper. We conduct the same experiments on two other models, \AVOD and \vPIXOR, and observe similar trends of improvements brought by learning with the depth loss (from \PSMNet to \PSMNet+DL), constructing the depth cost volume (from \PSMNet+DL to \SDN), and applying \GDC to correct the bias in stereo depth estimation (comparing \SDN+\GDC with \SDN).

We note that, in \autoref{stbl:ablation_lidar}, results of \AVOD (or \vPIXOR) with \SDN + L\# are worse than those with L\# at the moderate and hard settings. This observation is different from that in \autoref{tbl:ablation_lidar}, where \PRCNN with \SDN + L\# outperforms \PRCNN with L\# in 5 out of 6 comparisons. We hypothesize that this is because \PRCNN takes sparsified inputs (see \autoref{ssec:3D}) while \AVOD and \vPIXOR take dense inputs. In the later case, the four replaced LiDAR beams in \SDN+ L\# will be dominated by the dense stereo depths so that \SDN+ L\# is worse than L\#.

\subsection{Using fewer LiDAR beams}
\label{ssec:FewerLiDAR}
In PL++ (\ie, \SDN+ \GDC), we use 4-beam LiDAR to correct the predicted point cloud. In \autoref{tbl:ablation_beam}, we investigate using fewer (and also potentially cheaper) LiDAR beams for depth correction. We observe that even with 2 beams, \GDC can already manage to combine the two signals and yield a better performance than using 2-beam LiDAR or pseudo-LiDAR alone.

\begin{table*}[!htb]
	\centering
	\caption{\textbf{Ablation study on stereo depth estimation.} We report \APBEV ~/ \AP (in \%) of the \textbf{car} category at IoU$=0.7$ on the KITTI validation set. DL stands for depth loss. \label{stbl:ablation_sdn}}
	\tabcolsep 2.5pt
	\begin{tabular}{l|c|c|c|c|c|c}
		\hline
		\multicolumn{1}{c|}{\multirow{2}{*}{Depth Estimation}}                                                           & \multicolumn{3}{c|}{\vPIXOR} & \multicolumn{3}{c}{\AVOD} \\ \cline{2-7}
		\multicolumn{1}{c|}{}                                                   & Easy        & Moderate   & Hard   & Easy        & Moderate   & Hard     \\ \hline
		\PSMNetpd                                                                  &  73.9 / - \hspace{10pt}   & 54.0 / - \hspace{10pt}  & 46.9 / -  \hspace{10pt} & 74.9 / 61.9 &  56.8 / 45.3 & 49.0 / 39.0  \\
		\PSMNetpd + DL                                                             &  75.8 / - \hspace{10pt}   & 56.2 / - \hspace{10pt}  & 51.9 / - \hspace{10pt}  & 75.7 / 60.5 & 57.1 / 44.8 & 49.2 / 38.4  \\
		\SDN    & 79.7 / - \hspace{10pt}   & 61.1 / - \hspace{10pt}  & 54.5 / - \hspace{10pt}  & 77.0 / 63.2 & 63.7 / 46.8 & 56.0 / 39.8  \\ \hline
	\end{tabular}
\end{table*}

\begin{table*}[!htb]
	\centering
	\caption{\textbf{Ablation study on leveraging sparse LiDAR.} We report \APBEV ~/ \AP (in \%) of the \textbf{car} category at IoU$=0.7$ on the KITTI validation set. L\# stands for 4-beam LiDAR signal. \SDN+L\# means we replace the depth of a portion of pseudo-LiDAR points (\ie, landmark pixels) by L\#. \label{stbl:ablation_lidar}}
	\tabcolsep 2.5pt
	\begin{tabular}{l|c|c|c|c|c|c}
		\hline
		\multicolumn{1}{c|}{\multirow{2}{*}{Depth Estimation}} &  \multicolumn{3}{c|}{\vPIXOR} & \multicolumn{3}{c}{\AVOD} \\ \cline{2-7}
		\multicolumn{1}{c|}{}  &  Easy        & Moderate   & Hard     & Easy        & Moderate   & Hard  \\ \hline
		\SDN    &  79.7 / - \hspace{10pt}   & 61.1 / - \hspace{10pt}  & 54.5 / - \hspace{10pt}  & 77.0 / 63.2 & 63.7 / 46.8 & 56.0 / 39.8  \\
		L\#    & 72.0 / - \hspace{10pt}  & 64.7 / - \hspace{10pt} & 63.6 / - \hspace{10pt}  & 77.0 / 62.1 & 68.8 / 54.7 & 67.1 / 53.0  \\
		\SDN + L\#   & 75.6 / - \hspace{10pt}  & 59.4 / - \hspace{10pt} & 53.2 / - \hspace{10pt}  & 84.1 / 66.0 & 67.0 / 53.1 & 58.8 / 46.4  \\
		\SDN + \GDC    & 84.0 / - \hspace{10pt}  & 71.0 / - \hspace{10pt} & 65.2 / - \hspace{10pt}  & 86.8 / 70.7 & 76.6 / 56.2 & 68.7 / 53.4  \\ \hline
	\end{tabular}
\end{table*}

\begin{table*}[!htb]
	\centering
	\caption{\textbf{Ablation study on the sparsity of LiDAR.} We report \APBEV ~/ \AP (in \%) of the \textbf{car} category at IoU$=0.7$ on the KITTI validation set. L\# stands for using sparse LiDAR signal alone. The number in brackets indicates the number of beams in use. \label{tbl:ablation_beam}}
	\tabcolsep 2.5pt
	\begin{tabular}{l|c|c|c|c|c|c}
		\hline
		\multicolumn{1}{c|}{\multirow{2}{*}{Depth Estimation}} & \multicolumn{3}{c|}{\PRCNN}                                                            & \multicolumn{3}{c}{\vPIXOR} \\ \cline{2-7}
		\multicolumn{1}{c|}{}                                                   & Easy                            & Moderate                        & Hard                             & Easy        & Moderate   & Hard  \\ \hline
		SDN & 82.0 / 67.9 & 64.0 / 50.1 & 57.3 / 45.3 & 79.7 / - \hspace{10pt} & 61.1 / - \hspace{10pt} & 54.5 / - \hspace{10pt}  \\ \hline
		L\# (2)  & 69.2 / 46.3 & 62.8 / 41.9 & 61.3 / 40.0 & 66.8 / - \hspace{10pt}  & 55.5 / - \hspace{10pt} & 53.3 / - \hspace{10pt}    \\
		L\# (4)  & 73.2 / 56.1 & 71.3 / 53.1 & 70.5 / 51.5 & 72.0 / - \hspace{10pt}  & 64.7 / - \hspace{10pt} & 63.6 / - \hspace{10pt}    \\
		\hline
		\SDN + \GDC (2)    & 87.2 / 73.3 & 72.0 / 56.6 & 67.1 / 54.1 & 82.0 / - \hspace{10pt}  & 65.3 / - \hspace{10pt} & 61.7 / - \hspace{10pt}  \\
		\SDN + \GDC (4)   & 88.2 / 75.1 & 76.9 / 63.8 & 73.4 / 57.4 & 84.0 / - \hspace{10pt}  & 71.0 / - \hspace{10pt} & 65.2 / - \hspace{10pt}  \\
		\hline
	\end{tabular}
\end{table*}

\begin{table}[!h]
	\centering
	\tabcolsep 2.5pt
	\caption{\textbf{Comparison of \GDC and \PnP for 3D object detection.}  We report \APBEV ~/ \AP (in \%) of the \textbf{car} category at IoU$=0.7$ on the KITTI validation set, using \SDN+ \PnP or \SDN+ \GDC for depth estimation and \PRCNN or \vPIXOR for detection.} \label{tbDC}
	\begin{tabular}{l|c|c|c|c|c|c}
		\hline
		\multicolumn{1}{c|}{\multirow{2}{*}{Input signal}} & \multicolumn{3}{c|}{\PRCNN}                                                            & \multicolumn{3}{c}{\vPIXOR} \\ \cline{2-7}
		\multicolumn{1}{c|}{}                                                   & Easy                            & Moderate                        & Hard                             & Easy        & Moderate   & Hard  \\ \hline
		\SDN+ \PnP    & 86.3 / 72.1 & 73.3 / 58.9 & 67.2 / 54.2 & 79.1 / - \hspace{10pt}  & 64.2  / - \hspace{10pt} & 54.0 / - \hspace{10pt}  \\
		\SDN+ \GDC   & 88.2 / 75.1 & 76.9 / 63.8 & 73.4 / 57.4 & 84.0 / - \hspace{10pt}  & 71.0 / - \hspace{10pt} & 65.2 / - \hspace{10pt}  \\
		\hline
	\end{tabular}
\end{table}

\subsection{Depth correction vs. depth completion}
\begin{figure}
	\centering
	\begin{minipage}[t]{.53\textwidth}
		\centering
		\includegraphics[width=\textwidth]{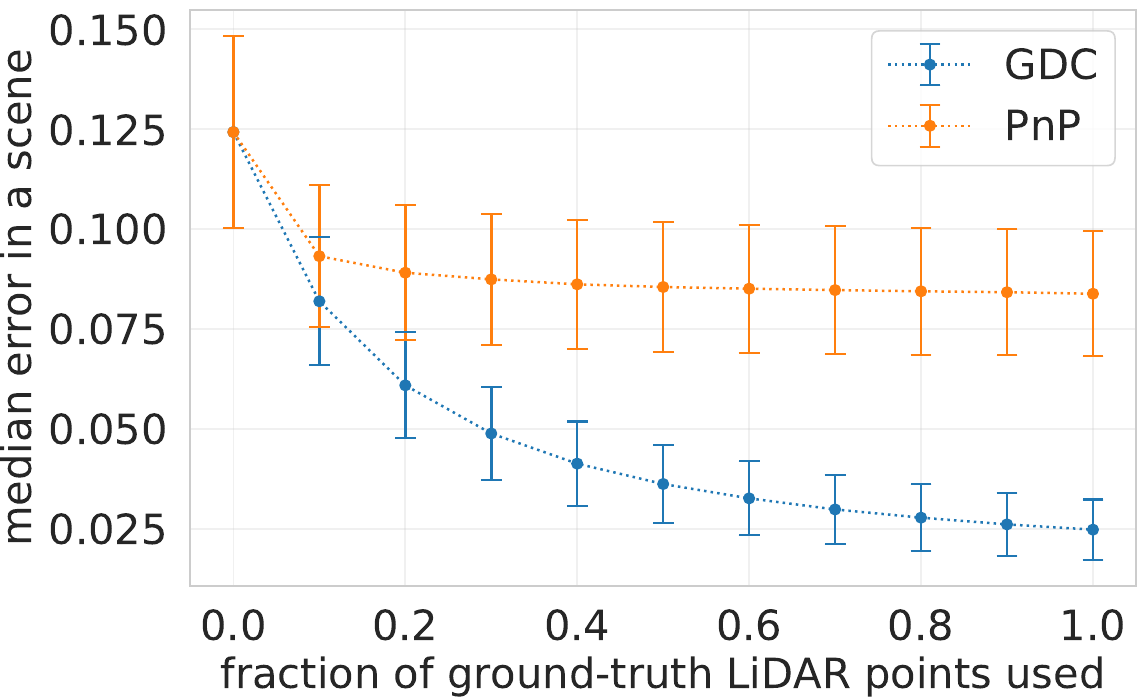}
		\caption{\textbf{Comparison of \GDC and \PnP for depth correction.} We report the median of absolute errors on the KITTI validation set. See text for details. \label{fig:depth_correction}}
	\end{minipage}
	\hfill
	\begin{minipage}[t]{.43\textwidth}
		\centering
		\includegraphics[width=\textwidth]{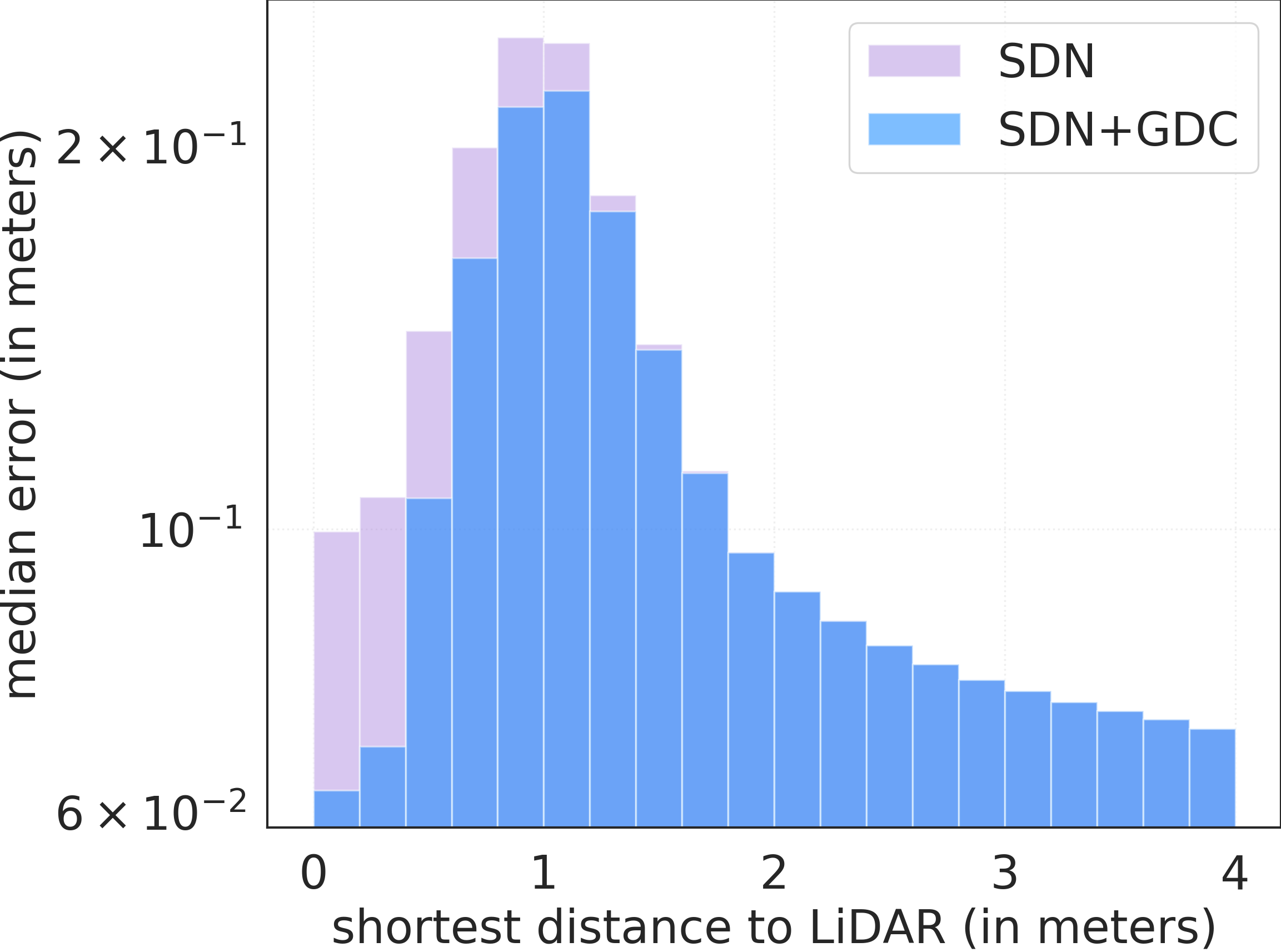}
		\caption{\textbf{Median depth estimation errors w.r.t. the shortest distances to 4-beam LiDAR points on KITTI validation set.} \label{fig:DepthValue_shortdist}}
	\end{minipage}
\end{figure}
We compare our \GDC algorithm for depth correction to depth completion algorithms, which aim to ``densify'' LiDAR data beyond the beam lines~\citep{wang2018plug,tomasello2018dscnet,ma2019self,yang2019dense,cheng2018depth, torres2004statistical}\footnote{\cite{torres2004statistical} use MRFs and may thus require less (or even no) training data compared to deep learning algorithms: a property shared by GDC.}. We note that most depth completion approaches take as input a 64-beam LiDAR and a single image, while our focus is on fusing a much sparser 4-beam LiDAR and stereo depths. As such, the two problems are not commensurate. Also, our \GDC algorithm is a general, simple, inference-time approach that \emph{requires no training}, unlike prior learning-based approaches to depth completion.

Here we empirically compare to \PnP~\citep{wang2018plug}, a recently proposed depth completion algorithm compatible with any (even stereo) depth estimation network, similar to \GDC. We use \SDN for initial depth estimation, and evaluate \GDC and \PnP by randomly selecting a fraction of LiDAR points as provided ground truths and calculating the median absolute depth errors on the remaining LiDAR points.
As shown in \autoref{fig:depth_correction}, \GDC outperforms \PnP by a large margin.
\autoref{tbDC} shows a further comparison to \PnP on 3D object detection. We apply \PnP and \GDC respectively to correct the depth estimates obtained from \SDN, train a \PRCNN or \vPIXOR using the resulting pseudo-LiDAR points on the KITTI training set, and compare the detection results on the KITTI validation set. In either case, \SDN + \GDC outperforms \SDN + \PnP by a notable margin.

\begin{table}
	\centering
	\caption{\textbf{Comparison of 3D object detection using the naive and optimized implementation of \GDC.} We report \APBEV ~/ \AP (in \%) of the \textbf{car} category at IoU$=0.7$ on the KITTI validation set, using \PRCNN for detection.} \label{tbOPT}
	\begin{tabular}{=l|+c|+c|+c}
		& Easy & Moderate & Hard \\ \hline
		Naive & 88.2 / 75.1 & 76.9 / 63.8 & 73.4 / 57.4 \\
		Optimized & 87.6 / 75.0 & 76.3 / 63.4 & 73.1 / 57.0\\
		\hline
	\end{tabular}
\end{table}

\subsection{Run time}
With the following optimizations for implementation,
\begin{enumerate}
	\item Sub-sampling pseudo-LiDAR points: keeping at most one point within a cubic of size $0.1\text{m}^3$
	\item Limiting the pseudo-LiDAR points for depth correction: keeping only those whose elevation angles are within $[-3.0^{\circ}, 0.4^{\circ})$ (the range of 4-beam LiDAR plus $0.6^{\circ}$; see \autoref{ssec:sparse} for details)
	\item After performing \GDC for depth correction, combining the corrected pseudo-LiDAR points with those outsides the elevation angles of $[-3.0^{\circ}, 0.4^{\circ})$
\end{enumerate}
\GDC runs in $90$ ms/frame using a single GPU ($7.7$ms for KD-tree construction and search, $46.5$ms for solving $W$, and $26.9$ms for solving $Z_{PL}'$) with negligible performance difference (see \autoref{tbOPT}). For consistency, all results reported in the main paper are based on the naive implementation.
Further speedups can be achieved by CUDA programming for GPUs.

\subsection{Stereo depth vs. detection}
\label{sec::supp_stereo_error}

We quantitatively evaluate the stereo depths by median errors in \autoref{fig:depth_comp} of the main text (numerical values are listed in \autoref{tbDepthValue}). In \autoref{tbDepthValue_mean} we further show mean errors with standard deviation (the large standard deviation likely results from outliers such as occluded pixels around object boundaries). For both tables, we divide pixels into beams according to their truth depths, and evaluate on pixels not on the 4-beam LiDAR. The improvement of \SDN (+ \GDC) over \PSMNetpd becomes larger as we consider pixels farther away. \autoref{tbRange} further demonstrates the relationship between depth quality and detection accuracy: \SDN (+ \GDC) significantly outperforms \PSMNetpd for detecting faraway cars. We note that, for very faraway cars (\ie, 50-70 m), the number of training object instances are extremely small, which suggests that the very poor performance might partially cause by over-fitting.

Further, we apply the same evaluation procedure but group the errors by the shortest distance between each \PL point and the 4-beam LiDAR points in \autoref{fig:DepthValue_shortdist}. We can see that the closer the \PL points are to the 4-beam LiDAR points, the bigger improvement GDC can bring.

\begin{table}[t]
	\centering
	\caption{\textbf{Median depth estimation errors over various depth ranges (numerical values of \autoref{fig:depth_comp}).} \label{tbDepthValue}}
	\begin{tabular}{l|c|c|c|c|c|c|c}
		\multicolumn{1}{c|}{\multirow{2}{*}{Signal}} & \multicolumn{7}{c}{range (m)}                                                                                                                                                                         \\ \cline{2-8}
		\multicolumn{1}{c|}{}                        & \multicolumn{1}{c|}{0-10} & \multicolumn{1}{c|}{10-20} & \multicolumn{1}{c|}{20-30} & \multicolumn{1}{c|}{30-40} & \multicolumn{1}{c|}{40-50} & \multicolumn{1}{c|}{50-60} & \multicolumn{1}{c}{60-70} \\ \hline
		PSMNet                       & 0.04                      & 0.11                       & 0.36                       & 0.83                       & 1.24                       & 1.98                       & 2.43                      \\
		SDN                                          & 0.07                      & 0.12                       & 0.30                       & 0.60                       & 0.89                       & 1.31                       & 1.73                      \\
		SDN + GDC                                    & 0.07                      & 0.12                       & 0.27                       & 0.51                       & 0.74                       & 1.03                       & 1.53                      \\ \hline
	\end{tabular}
\end{table}
\begin{table}[t]
	\centering
	\caption{\textbf{Mean depth estimation errors (with standard deviation) over various depth ranges.} \label{tbDepthValue_mean}}
	\tabcolsep 2pt
	\begin{tabular}{l|c|c|c|c|c|c|c}
		\multicolumn{1}{c|}{\multirow{2}{*}{Signal}} & \multicolumn{7}{c}{range (m)}                                                                                                                                                                         \\ \cline{2-8}
		\multicolumn{1}{c|}{}                        & \multicolumn{1}{c|}{0-10} & \multicolumn{1}{c|}{10-20} & \multicolumn{1}{c|}{20-30} & \multicolumn{1}{c|}{30-40} & \multicolumn{1}{c|}{40-50} & \multicolumn{1}{c|}{50-60} & \multicolumn{1}{c}{60-70} \\ \hline
		PSMNet                       & 0.18$\pm$0.93                      & 0.36$\pm$1.20                     & 0.97$\pm$2.32                     & 2.02$\pm$4.05                     & 2.94$\pm$5.64                     & 4.61$\pm$8.03                     & 6.03$\pm$10.32                    \\
		SDN                                          & 0.21$\pm$0.89                    & 0.35$\pm$1.16                     & 0.87$\pm$2.31                     & 1.80$\pm$4.22                     & 2.67$\pm$6.00                     & 4.27$\pm$8.78                     & 5.82$\pm$11.23                    \\
		SDN + GDC                                    & 0.21$\pm$0.90                    & 0.35$\pm$1.17                     & 0.84$\pm$2.34                     & 1.74$\pm$4.27                     & 2.59$\pm$6.06                     & 4.14$\pm$8.85                     & 5.72$\pm$11.29                    \\ \hline
	\end{tabular}
\end{table}

\begin{table}[t]
	\centering
	\caption{\textbf{3D object detection at various depth ranges.} We compare different input signals. We report \APBEV ~/ \AP (in \%) of the \textbf{car} category at IoU$=0.7$ on the KITTI validation set, using \PRCNN for detection. In the last two rows we show the number of car objects in KITTI object train and validation sets within different ranges.} \label{tbRange}
	\vskip-5pt
	\begin{tabular}{=l|+c|+c|+c}
		Input signal & 0-30 m & 30-50 m & 50-70 m \\ \hline
		\PSMNet & 65.6 / 54.0 & 15.8 / \hspace{2pt} 6.9  &  \hspace{2pt} 0.0 / \hspace{2pt} 0.0 \\
		\SDN & 68.6 / 56.7 & 27.4 / 11.3 & \hspace{2pt} 0.7 / \hspace{2pt} 0.0 \\
		\SDN+ \GDC & 84.7 / 67.8 & 49.9 / 31.5 & \hspace{2pt} 2.5 / \hspace{2pt} 1.0\\
		\rowstyle{\color{gray}}
		\textsc{LiDAR} & 88.5 / 84.0 & 69.9 / 51.5 & \hspace{2pt} 8.9 / \hspace{2pt} 3.4 \\ \hhline{====}
		\textsc{\#~Objects-Train} & 6903 & 3768 & 76 \\
		\textsc{\#~Objects-Val} & 7379 & 3542 & 39 \\ \hline
	\end{tabular}
	\vspace{-10pt}
\end{table}

\subsection{Connected Components in KNN graphs of \PL points by SDN}
\label{sec::supp_connected_components}
\begin{figure}
	\centering
	\begin{minipage}[t]{.48\textwidth}
		\centering
		\includegraphics[width=.9\linewidth]{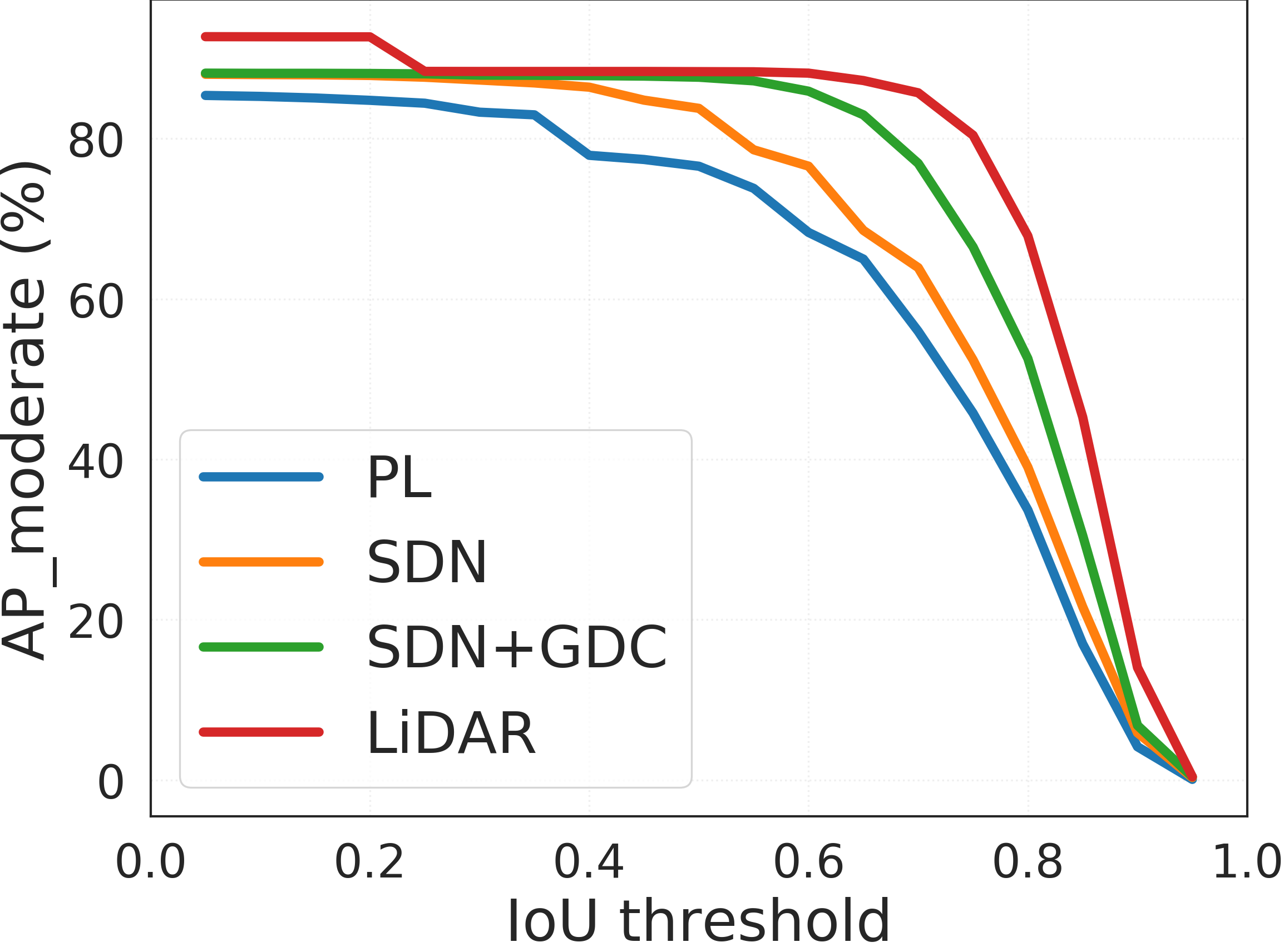}
		\caption{\textbf{IoU vs. \APBEV on KITTI validation set on the car category (moderate).}\label{fig:IoU}}
	\end{minipage}
	\hfill
	\begin{minipage}[t]{.48\textwidth}
		\centering
		\includegraphics[width=.9\linewidth]{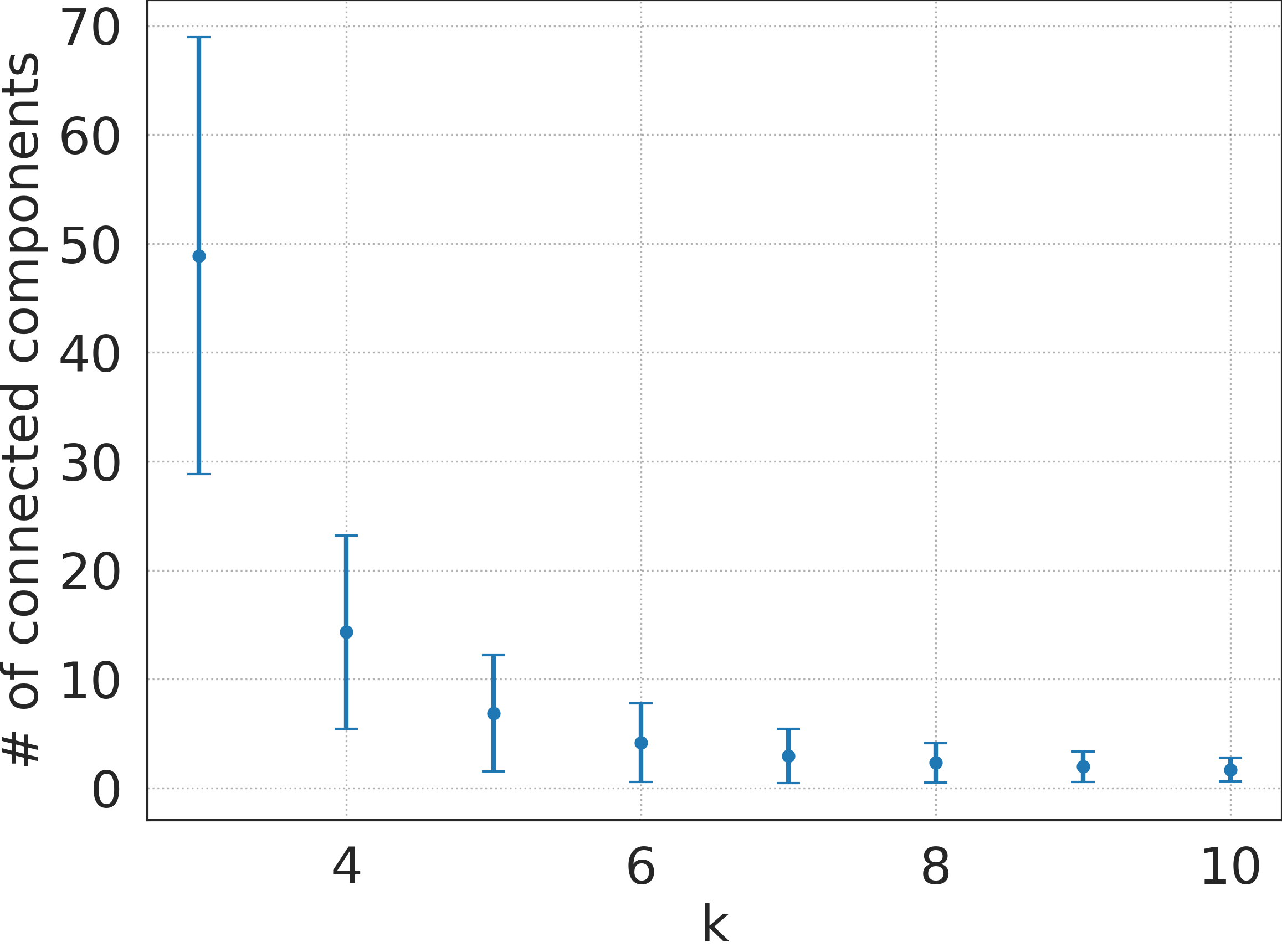}
		\caption{\textbf{$k$ vs. average number of connected components in KNN graphs of \PL points by SDN.}\label{fig:cc}}
	\end{minipage}

\end{figure}

Here, we provide empirical analysis on the relationship between the $k$ we choose in building the K-nearest-neighbor graph of \PL points by SDN and the number of connected components of that graph. We show the results on KITTI validation set in \autoref{fig:cc}. It can be seen that with $k \geq 9$, the average number of connected components in the graph is smaller than $2$.

\subsection{Failure cases and weakness}
There is still a gap between our approach and LiDAR for faraway objects (see \autoref{tbRange}). We further analyze \APBEV at different IoU in \autoref{fig:IoU}. For low IoU (0.2-0.5), \SDN (+\GDC) is on par with LiDAR, but the gap increases significantly at high IoU thresholds.
This suggests that the predominant gap between our approach and LiDAR is because of mislocalization, perhaps due to residual inaccuracies in depth.

\subsection{Qualitative results}
\label{ssec:QR}
\begin{figure*}[htb!]
	\centering
	\includegraphics[width=.85\linewidth]{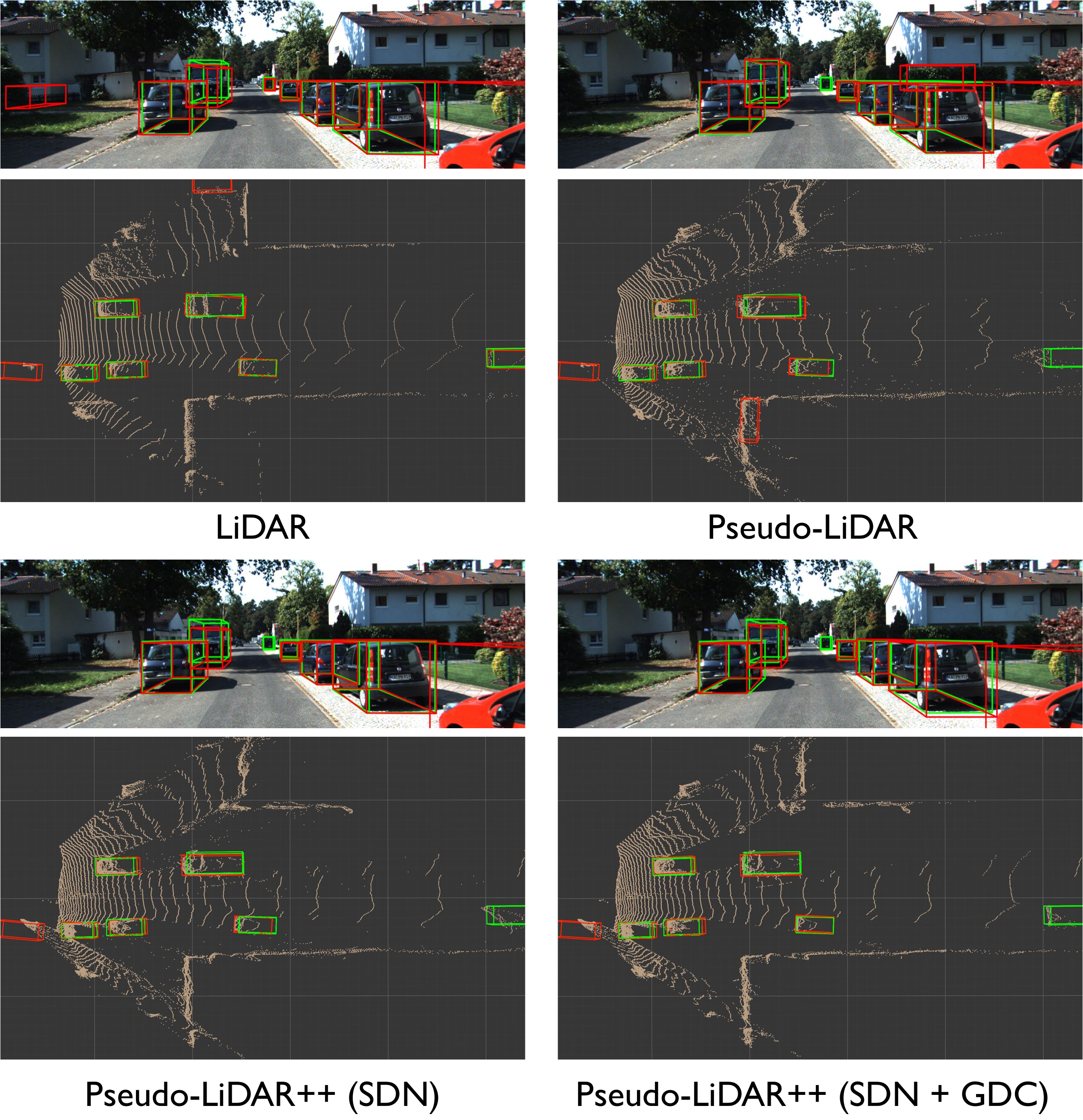}
	\caption{\textbf{Qualitative Comparison.} We show the detection results on a KITTI validation scene by \PRCNN with different input point clouds. We visualize them from both frontal-view images and bird's-eye view (BEV) point maps. Ground-truth boxes are in green and predicted bounding boxes are in red. The observer is at the left-hand side of the BEV map looking to the right. In other words, ground truth boxes on the right are more faraway (\ie, deeper) from the observer, and hence hard to localize. Best viewed in color. \label{fig:qualitative_2}}
\end{figure*}

\begin{figure*}[htb!]
	\centering
	\includegraphics[width=.85\linewidth]{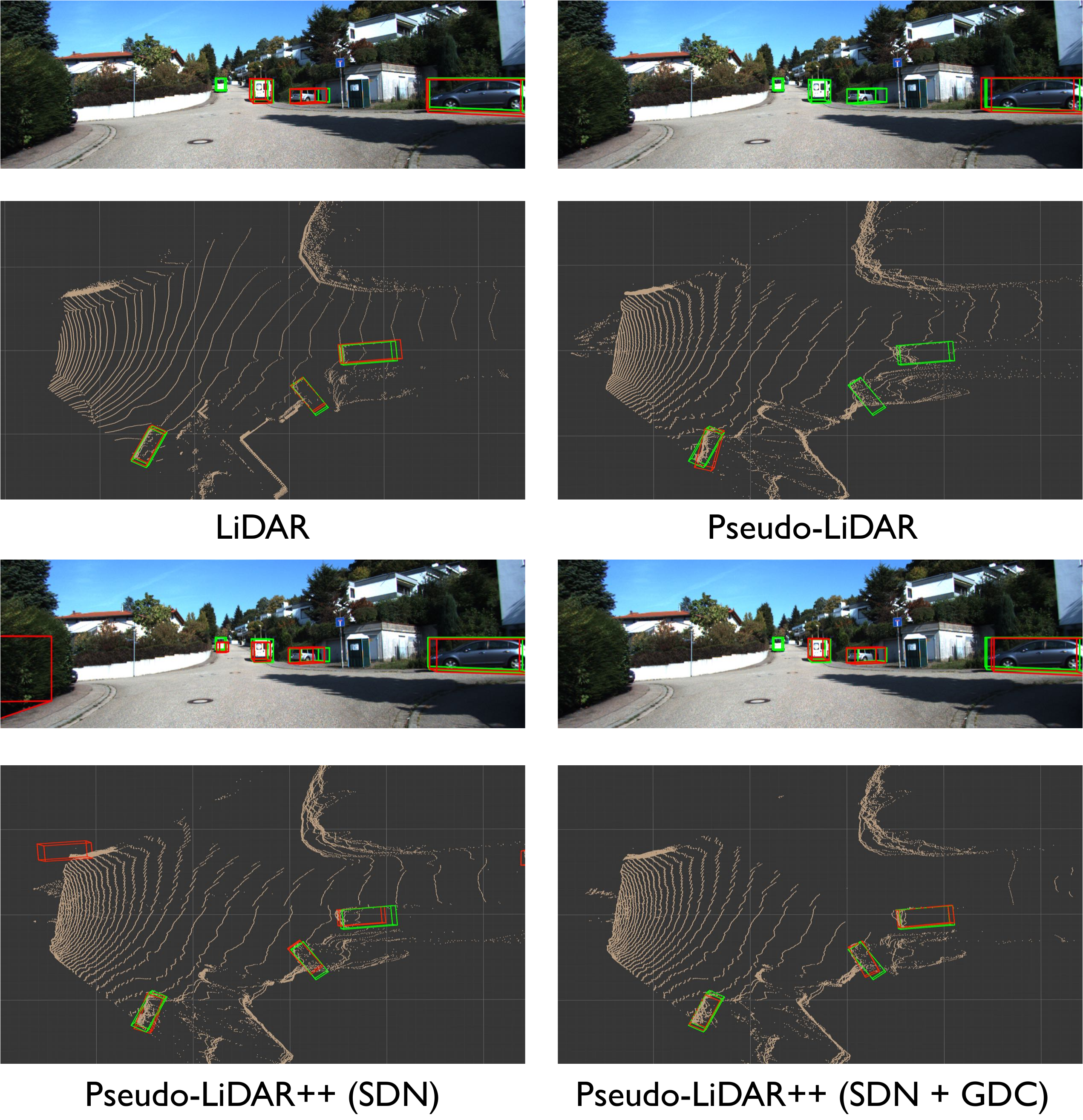}
	\caption{\textbf{Qualitative Comparison - another example.} The same setup as in \autoref{fig:qualitative_2} \label{fig:qualitative_3}}
\end{figure*}

\begin{figure*}[htb!]
	\centering
	\includegraphics[width=.85\linewidth]{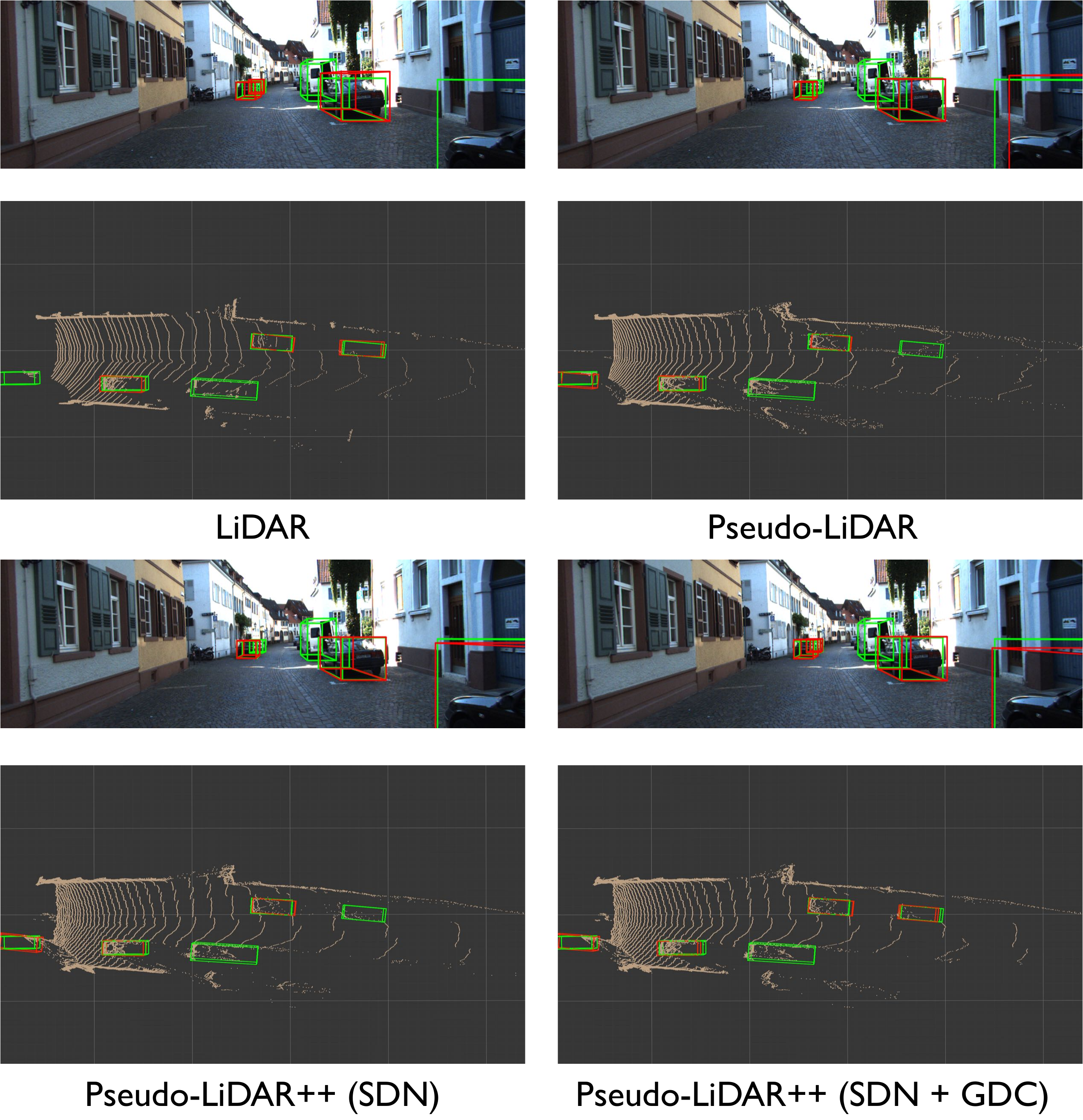}
	\caption{\textbf{Qualitative Comparison - another example.} The same setup as in \autoref{fig:qualitative_2} \label{fig:qualitative_4}}
\end{figure*}

In Figure~\ref{fig:qualitative},\ref{fig:qualitative_2},\ref{fig:qualitative_3} and \autoref{fig:qualitative_4}, we show detection results using \PRCNN (with confidence $> 0.95$) with different input signals on four randomly chosen scenes in the KITTI object validation set. Specifically, we show the results from the frontal-view images and the bird's-eye view (BEV) point clouds. In the BEV map, the observer is on the left-hand side looking to the right. It can be seen that the point clouds generated by \PL++ (\SDN alone or \SDN+\GDC) align better with LiDAR than those generated by \PL (\PSMNet).
For nearby objects (\ie, bounding boxes close to the left in the BEV map), we see that \PRCNN with any point cloud performs fairly well in localization. However, for faraway objects (\ie, bounding boxes close to the right), \PL with depth estimated from \PSMNet predicts objects (red boxes) deviated from the ground truths (green boxes). Moreover, the noisy \PSMNet points also leads to several false positives or negatives. In contrast, the detected boxes by our \PL++, either with \SDN alone or with \SDN+\GDC, align pretty well with the ground truth boxes, justifying our targeted improvement in estimating faraway depths. In \autoref{fig:qualitative_2}, we see one failure case for both \PL and \PL++: the most faraway car is missed, while LiDAR signal can still detect it, suggesting that for very faraway objects stereo-based methods may still have limitation.

\end{document}